\documentclass[acmsmall, manuscript,screen,authorversion,nonacm]{acmart}
\usepackage{tikz}
\usepackage{ifthen}
\usetikzlibrary{matrix,calc}
\usepackage{multirow}
\usepackage{algorithm} 
\usepackage{algpseudocode} 
\usepackage{graphicx}
\usepackage{multicol}

\AtBeginDocument{%
  \providecommand\BibTeX{{%
    \normalfont B\kern-0.5em{\scshape i\kern-0.25em b}\kern-0.8em\TeX}}}
\setcopyright{acmcopyright}
\copyrightyear{2022}
\acmYear{2022}
\acmConference[CSCW '22]{CSCW '22: Proceedings of the ACM Human and Computer Interaction}{January 15, 2022}{Cambridge, MA}
\acmBooktitle{CSCW '22: Proceedings of the ACM Human and Computer Interaction, January 15, 2022, Cambridge, MA}
\acmPrice{15.00}
\acmISBN{978-1-4503-XXXX-X/18/06}
\begin{document}
\title[Towards Transparency in Dermatology]{Towards Transparency in Dermatology Image Datasets with Skin Tone Annotations by Experts, Crowds, and an Algorithm}
\author{Matthew Groh}
\email{groh@mit.edu}
\orcid{0000-0002-9029-0157}
\affiliation{%
  \institution{MIT Media Lab}
  \streetaddress{75 Amherst Street}
  \city{Cambridge}
  \state{MA}
  \country{USA}
  \postcode{02139}
}
\author{Caleb Harris}
\affiliation{%
  \institution{MIT Media Lab} \country{USA}}
\email{harriscw@mit.edu}
\author{Roxana Daneshjou}
\affiliation{%
  \institution{Stanford University} \country{USA}}
\email{roxanad@stanford.edu}
\author{Omar Badri}
\affiliation{%
  \institution{Northeast Dermatology Associates} \country{USA}}
\email{obadri@gmail.com}
\author{Arash Koochek}
\affiliation{%
  \institution{Banner Health} \country{USA}}
\email{arash.koochek@bannerhealth.com}
\renewcommand{\shortauthors}{Groh, et al.}
\begin{abstract}
  While artificial intelligence (AI) holds promise for supporting healthcare providers and improving the accuracy of medical diagnoses, a lack of transparency in the composition of datasets exposes AI models to the possibility of unintentional and avoidable mistakes. In particular, public and private image datasets of dermatological conditions rarely include information on skin color. As a start towards increasing transparency, AI researchers have appropriated the use of the Fitzpatrick skin type (FST) from a measure of patient photosensitivity to a measure for estimating skin tone in algorithmic audits of computer vision applications including facial recognition and dermatology diagnosis. In order to understand the variability of estimated FST annotations on images, we compare several FST annotation methods on a diverse set of 460 images of skin conditions from both textbooks and online dermatology atlases. These methods include expert annotation by board-certified dermatologists, algorithmic annotation via the Individual Typology Angle algorithm, which is then converted to estimated FST (ITA-FST), and two crowd-sourced, dynamic consensus protocols for annotating estimated FSTs. We find the inter-rater reliability between three board-certified dermatologists is comparable to the inter-rater reliability between the board-certified dermatologists and either of the crowdsourcing methods. In contrast, we find that the ITA-FST method produces annotations that are significantly less correlated with the experts' annotations than the experts' annotations are correlated with each other. These results demonstrate that algorithms based on ITA-FST are not reliable for annotating large-scale image datasets, but human-centered, crowd-based protocols can reliably add skin type transparency to dermatology datasets. Furthermore, we introduce the concept of dynamic consensus protocols with tunable parameters including expert review that increase the visibility of crowdwork and provide guidance for future crowdsourced annotations of large image datasets.

\end{abstract}
\maketitle
\section{Introduction}

Artificial intelligence (AI) algorithms hold promise for improving image-based clinical diagnosis tasks ranging from identifying breast cancer in mammograms~\cite{mckinney_international_2020} to classifying skin lesions based on a single image~\cite{esteva_dermatologist-level_2017} to predicting the diagnosis of hundreds of diverse skin conditions based on a few images and a brief patient history~\cite{liu_deep_2020}. The combination of algorithmic predictions with physician diagnostic skill has the potential to create large efficiency and welfare gains in healthcare~\cite{ribers_machine_2020}. In particular, AI systems can enhance specialists' diagnostic performance on specific tasks (e.g. identifying pneumonia on chest radiographs~\cite{patel2019human} and predicting hypoxaemia risk from operating room data~\cite{lundberg2018explainable}) but incorrect predictions from an AI system can mislead specialists and generalists alike~\cite{tschandl_humancomputer_2020, jacobs2021machine, groh2022deepfake}. In fact, inaccurate advice regardless of whether it comes from an AI or human tends to decrease physicians' accuracy on diagnostic tasks~\cite{gaube_as_2021}. Moreover, the algorithm appreciation effect~\cite{logg2019algorithm} suggests that inaccurate advice from an algorithm is likely to have more negative effects than the same advice given by a human. 

Given the consequences of inaccurate advice in healthcare, ethical and responsible algorithm-in-the-loop decision systems should require the systems to be both accurate and also unbiased with regard to sensitive attributes like race and gender. Moreover, these systems should be transparent such that medical experts can reliably assess algorithmic performance~\cite{green2019principles, haibe2020transparency}. These principles for ethical systems are particularly important because algorithms are prone to make unexpected errors on out-of-distribution data. Due to biases in dataset representation, protected classes are more likely to be out-of-distribution~\cite{barocas_big_2016,alkhatib2019street,liu_understanding_2021}. Moreover, when accurate yet non-equitable algorithmic risk assessments are used as decision support tools they have been shown to alter decision maker's risk aversion and lead to unexpected and sometimes unwanted shifts in human decision-making~\cite{green2021algorithmic}.

Yet the vast majority of AI algorithms for diagnostic tasks in dermatology are trained on datasets that lack transparency with regards to demographic and skin tone attributes~\cite{daneshjou_lack_2021, wen2021}. Due to this lack of transparency, it is difficult to assess what data may be out-of-distribution and this leads to the potential for unexpected errors that could have otherwise been addressed. For example, given the under representation of dark skin in educational resources~\cite{ebede_disparities_2006,alvarado_representation_2020,adelekun_skin_2020, louie_representations_2018, lester_absence_2020} and online dermatology atlases~\cite{groh_evaluating_2021}, it is unknown the full extent to which dark skin is under-represented in many of the large dermatology image datasets. For the few datasets that do include skin type information, dark skin types are underrepresented ~\cite{daneshjou_lack_2021}. This is particularly problematic because AI algorithms for classifying the skin condition in an image are more accurate on images that match the skin color upon which the algorithm is trained than images that do not match the skin color~\cite{groh_evaluating_2021}. An analysis of three AI algorithms (ModelDerm~\cite{han2020augmented}, DeepDerm~\cite{esteva_dermatologist-level_2017}, and Ham 1000~\cite{tschandl2018ham10000}) reveal that images of dark skin show a drop in all three models' accuracy rates relative to rates in images of light skin~\cite{daneshjou2021disparities}.

One approach for increasing transparency in dermatology image datasets and their resulting AI algorithms is to annotate skin tone with Fitzpatrick Skin Type (FST) like the algorithmic audit of accuracy disparities in facial recognition by Buolamwini and Gebru 2018~\cite{buolamwini2018gender}. FST is a clinical measurement developed and used by dermatologists to assess patients' sun sensitivity for dosing phototherapy or chemophototherapy. Clinical FST has been criticized for subjectivity~\cite{gupta2019skin}, is not designed for classifying race or skin color~\cite{ware_racial_nodate}, and often involves not just assessing skin tone but assessing a patient's hair color and eye color~\cite{noauthor_fitzpatrick_2015}. Despite the imperfections and biases of clinical FST as a proxy for skin tone and an assessment of differential healthcare risks~\cite{okoji2021equity}, AI researchers have appropriated FST to estimate skin tone labels for algorithmic audits of tasks like classifying skin disease~\cite{liu_deep_2020,dulmage_point--care_2020, phillips2019assessment, groh_evaluating_2021} and facial recognition algorithms~\cite{buolamwini2018gender, hazirbas_towards_nodate}. In this paper, we distinguish FST as recorded in a clinical patient-provider interaction as ``clinical FST'' and FST as recorded based on a single image as ``estimated FST.''

While estimated FST has been frequently used in computer vision tasks, basic questions have not been explored about its use for labeling image datasets: Who is qualified to annotate images with estimated FSTs? More specifically, should estimated FST annotations on large-scale datasets be limited to board-certified dermatologists? How concordant are board-certified dermatologists, particularly on the kinds of datasets used in computer vision? Would the annotations of trained annotators, crowdsourced labor, or algorithms differ significantly from board-certified dermatologists? These are empirical questions, which are connected more broadly to questions about what makes desirable data and how race and gender should be annotated in image datasets~\cite{scheuerman_how_2020, scheuerman_datasets_2021}. Notably, we limit the focus on estimated FST because it is a method used in algorithmic audits based on clinical medicine and it allows granular analysis which would not be captured by race alone~\cite{buolamwini2018gender}. While most large image datasets with estimated FST annotations are labeled by dermatologists~\cite{liu_deep_2020, dulmage_point--care_2020, phillips2019assessment, buolamwini2018gender}, the ``Casual Conversations'' dataset is annotated by trained annotators~\cite{hazirbas_towards_nodate} and the ``Fitzpatrick 17k'' annotations are generated by applying a dynamic consensus protocol to crowdsourced annotations~\cite{groh_evaluating_2021}. 

As an alternative to human-annotated estimated FST, researchers have proposed and used the Individual Typology Angle to FST (ITA-FST) algorithm, a computer vision algorithm that converts the RGB values of an image into a single metric for constitutive pigmentation, to estimate apparent skin tone from images~\cite{kinyanjui_estimating_2019,krishnapriya2022analysis}. Prior work shows that ITA-FST is strongly correlated with Melanin Index~\cite{noauthor_fitzpatrick_2015}, which is sometimes used in assigning clinical FSTs~\cite{eilers2013accuracy}. However, recent research in photodermatology suggests that ITA used for constitutive pigmentation is a poor proxy for clinical FST~\cite{osto2022individual}.    

Prior work suggests that crowdsourced estimated FST annotations are generally within 1 point of an expert board-certified dermatologist's annotation, but Groh et al (2021) compared crowd annotations with only a single expert, do not include statistical analyses of inter-rater reliability, do not compare ITA with experts' annotations, and do not examine nuances around the compositions of the crowd or edge cases where the crowd is prone to err~\cite{groh_evaluating_2021}. We present evidence that the inter-rater reliability between three board-certified dermatologists is comparable to the inter-rater reliability between board-certified dermatologists and crowdsourcing methods but not the ITA-FST algorithm. However, for a subset of images with high disagreement between crowd annotators, we find higher inter-rater reliability between board-certified dermatologists than board-certified dermatologists and the crowd. 

In summary, our contributions are the following:

\quad\textbf{(1)} We evaluate the inter-rater reliability between three medical experts, a computer vision algorithm, and two crowdsourcing approaches for annotating images of skin conditions with estimated FST, which is useful for increasing transparency into how algorithms perform on images of different skin tones. On a set of 320 images drawn from dermatology textbooks~\cite{jackson2014dermatology,kailas2017taylor,kang2019fitzpatrick,wolff2008fitzpatricks,habif2010clinical,bolognia2018dermatologia,griffiths2016rook,micheletti2022andrews}, we do not find a statistically significant difference when comparing the Pearson Correlation Coefficients ($\rho$) between three medical experts with the $\rho$ between each medical expert and either of the crowdsourcing methods. In contrast, we do find a statistically significant difference in the annotations produced by the ITA-FST algorithm. These results suggest that crowdsourcing (but not the ITA-FST algorithm) can be a reliable source for generating estimated FST annotations on large-scale datasets of images intended for training and evaluating AI models to classify skin disease. However, we include important caveats. First, our qualitative results show the crowd will sometimes make errors that the medical experts would be unlikely to make. Second, a quantitative follow-up with 140 images drawn from two online dermatology atlases~\cite{alkattash, desilva} shows the results are robust to 70 images randomly sampled from the 91\% of images with relatively low crowd disagreement but on a random sample of 70 images from the 9\% of images with relatively high crowd disagreement,  we find the crowd annotations can be significantly different from the experts' annotations. Third, the image-based estimated FST annotations are subject to lighting, image quality, and pose variability that are not an issue for in-person assessments 

\quad\textbf{(2)} In order to increase visibility into the process of human annotation of large image datasets and guide future work, we introduce and describe a dynamic consensus protocol for aggregating crowdsourced estimated FST annotations using the following transparent, adjustable criteria: (1) \textbf{consensus thresholds}, (2) \textbf{qualified annotations}, (3) \textbf{failure reports}~\cite{cabrera_discovering_2021}, (4) \textbf{agreement metrics} and (5) \textbf{expert review}. We apply this procedure to the publicly available ``Fitzpatrick 17k'' dataset of 16,577 images to evaluate inter-rater reliability across crowdsourcing annotation methods, estimate the proportion of images that experts should review, and conduct expert review on 140 images.

\section{Background and Related Work}

\subsection{Data Documentation for Increasing Transparency and Accountability in Algorithms}

Critical frameworks documenting both machine learning datasets and their resulting models promote transparency and accountability by enabling nuanced analyses that can expose unwanted biases. Examples of guiding frameworks for detailing data, its definitions, and its associated models' potential harms include \textit{Data Statements for Natural Language Processing}, \textit{The Dataset Nutrition Label}~\cite{holland2018dataset}, \textit{Model Cards for Model Reporting}~\cite{mitchell2019model}, and \textit{Datasheets for Datasets}~\cite{gebru2021datasheets}. The seminal algorithmic audit of accuracy disparities in facial recognition by Buolamwini and Gebru 2018 relied on documenting estimated FST annotations and evaluating algorithmic performance across FSTs and found significant intersectional accuracy disparities~\cite{buolamwini2018gender}. Estimated FST annotations have also been helpful in documenting accuracy in machine learning models for dermatology~\cite{groh_evaluating_2021, daneshjou2021disparities}. With appropriate, inclusive data, algorithms can increase accountability by both serving as a diagnostic tool to detect discrimination and formalizing our definitions around a social problem like inequities in healthcare across gender, race, and skin color~\cite{abebe_roles_2020, kleinberg2020algorithms}. 

Beyond cataloguing the elements of a dataset, data documentation can also question the existence of categories within the data and inform the question posed by Miceli et al 2022: ``Is this information sufficient in itself to explicate unjust outcomes''~\cite{miceli2022studying}? For a large number of datasets with images of humans, the definitions of both race and gender in databases lack critical engagements, are overly reductive, and require more than an outside observer looking at a photograph to annotate appropriately~\cite{scheuerman_how_2020,miceli2021documenting}. This is particularly problematic because the definition of a category, class, or outcome will impact how disparate treatment and disparate impact arise in the data~\cite{barocas_big_2016}.   For example, Obermeyer et al 2019 report that an algorithm for predicting health risk of millions of people in the United States using cost of care as a proxy for health needs led to the following bias: ``At a given risk score, Black patients are considerably sicker than White patients, as evidenced by signs of uncontrolled illnesses''~\cite{obermeyer_dissecting_2019}. This racial bias in a healthcare setting is not only a problem of selecting the right outcome measure, but a deeper problem that involves a history in the United States of ``segregated hospital facilities, racist medical curricula, and unequal insurance structures, among other factors''~\cite{benjamin2019assessing}. Dataset documentation is an initial step that enables critical researchers to both identify empirical biases, question the definitions of specific data features, and inspect the data generating process. Data documentation is particularly helpful to bridge collaboration between data scientists and subject matter experts to make knowledge and processes explicit such that both groups of people can ask the right question~\cite{mao2019data}. As bias is uncovered in data, researchers can offer new insights into bias as a starting point for ``studying up''~\cite{nader1972up} with a critical focus on accountability and power dynamics in the underlying data generation process~\cite{barabas2020studying}.

Extracting categories and clusters from complex data involves value judgments. For example, Scheuerman et al 2021 highlight the tensions in the development of computer vision datasets between efficiency and care, universality and contextuality, impartiality and positionality, and model work and and data work~\cite{scheuerman_datasets_2021}. In crowd sourcing tasks, categorizing can become problematic when crowdworkers have limited attention and expertise~\cite{andre2014crowd} and when crowdworkers are overly constrained by power dynamics such that the crowd annotates data based on their expectations of how a client sees the world rather than their own sense of how the world looks~\cite{miceli2020between}. One recent applied example from CSCW shows that accessible interfaces with high degrees of freedom enable crowdworkers to categorize data that can appropriately filter harmful content generated by AI~\cite{mandel2020using}. Another example from recent research in CSCW highlights the potential for failure reports~\cite{cabrera_discovering_2021} – open-ended descriptions of model errors – to help navigate unexpected systematic failures. We expand on the concept of failure reports in Section 3.3 where crowdworkers can transcend the menu of multiple choice annotations (in our case estimated FST I-VI and ``not applicable'') to free-text responses where images can be flagged for being incorrectly labeled, inappropriate, or irrelevant. 

In dataset documentation, epistemic authority is an important value judgment. How should data be annotated and who or what should do it? Data annotation is often less straightforward and more complex than it seems. Data work is often time-consuming, opaque (unless there's good documentation), and not well rewarded; in field interviews with data workers, one interviewee exclaimed, ``Everyone wants to do the model work, not the data work,'' which is a sentiment shared by many interviewees~\cite{sambasivan2021everyone}. Moreover, reasonable people often disagree on color classification~\cite{goodwin2000practices} and medical experts often disagree on medical diagnoses~\cite{rajpurkar2017chexnet,raghu_direct_nodate}. In fact, in one study comparing referral and final diagnoses across 280 patients, significant disagreements appear in 21\% of cases~\cite{van2017extent}. Instead of assigning epistemic authority to any particular individual or algorithm for a subjective task, we follow prior work that treats epistemic authority based on inter-annotator agreement~\cite{cohen1960coefficient, fisher1915frequency, fisher1921probable, hotelling1953new}. Disagreement between annotators is not necessarily indicative of poor quality annotations or bias, but instead, disagreement can help reveal the subjectivity involved in a particular task and a particular example~\cite{aroyo2015truth}. 

In order to answer who or what is epistemically qualified to annotate data with information that can provide transparency and accountability into potential biases, we need to examine the level of agreement produced by different methods. The first step involves measuring the subjectivity of the task by measuring the degree of disagreement among experts. Next, we compare alternative methods (e.g. an algorithm and crowd methods) to the level of disagreement among experts. If an alternative method does not disagree significantly more with experts than experts do with each other, then we can call the alternative method generally comparable to experts. While an alternative method may be generally comparable to experts, edge cases may arise where experts have significantly lower disagreement among themselves than with the crowd. For example, in the case of estimated FST images where a rare skin disease has transformed the color of the skin, non-experts may have higher levels of disagreement than experts. In the framework of Muller et al 2019 \textit{How Data Science Workers Work with Data}~\cite{muller2019data}, this annotation process would be described as ``Ground Truth as Created'' where human expertise applied to images informs an analysis of the similarity of various methods for annotating estimated FST. While Muller et al 2021~\cite{muller2021designing} write that ``it is widely agreed that SME-labeled data [data labeled by subject matter experts] is the "gold standard" data source for high quality labeled data for specialized tasks,'' we take a step back from this assumption and empirically evaluate how well crowds and an algorithm compare to estimated FST annotations of images by board-certified dermatologists. The dynamic consensus threshold process described in 3.3 can represent both Muller et al 2021's ``Principled design'' and ``Iterative design'' to ground truth annotation because the annotation process is planned and well-defined, but aspects of the dynamic consensus threshold process (e.g. failure reports and expert review) allow for clarifications, adjustments, and potential re-definitions based on collaboration back and forth between the human annotators examining data at the record level and data scientists examining the data at the dataset level.

\subsection{Designing Transparency into Clinical Decision Support Systems}

Clinical decision support systems (CDSSs) are systems designed to support healthcare providers in medical decision-making. Past work at CSCW has documented the following relevant onboarding criteria for healthcare providers to interact with CDSSs: capabilities and limitations, functionality, design objective, relative strengths and weaknesses of an algorithm, performance of a model on domain specific cases including how the model's idiosyncrasies compare with human idiosyncrasies~\cite{cai_hello_2019,cai_onboarding_2021}. Transparency on subgroups within the data is an integral component to onboarding healthcare providers such that they develop an understanding for when they should override the system, which is important when an algorithm makes an erroneous prediction~\cite{de2020case}. In practice, healthcare professionals seek to compare algorithmic errors in CDSSs with their own errors~\cite{cai_human-centered_2019}. Well-documented data enables healthcare providers (and researchers) to examine subcategories on which algorithms are likely to error, which is important for establishing trust that the algorithm will lead to positive results for vulnerable patients~\cite{vereschak_how_2021, ehsan2021expanding} and useful for identifying what kind of data should be collected to reduce accuracy disparities~\cite{chen2018my}.

Recent deployments of deep learning systems for health reveal that algorithms trained on retrospective datasets may not be ecologically valid~\cite{beede2020human}. In particular, an algorithm applied to data that deviates from the training data is prone to unexpected errors on the out-of-distribution examples. One approach to handling out-of-distribution data is training a classifier to predict whether an image is out-of-distribution and if it is to abstain from generating an algorithmic classification~\cite{roy2022does}. Decision support systems tend to be less effective on out-of-distribution examples; in evaluations of algorithms that generally outperform humans, the performance gap in accuracy between humans informed by the algorithm and the algorithm is lower in out-of-distribution examples than in-distribution examples~\cite{liu_understanding_2021}. If a particular skin tone is out-of-distribution for a particular disease, this is important for clinicians and model developers to know, so they are aware what kind of data might constitute a context shift~\cite{contextshift}. 

Recently, Jain et al 2021 completed a retrospective study that showed how a deep learning based CDSS may help non-specialists such as primary care physicians and nurse practitioners diagnose skin disease with higher accuracy (defined as agreement with reference conditions) and possibly reduce biopsy and referral rates to dermatologists than the providers would without the system~\cite{jain_development_2021}. Decision support systems have the potential to improve the quality of dermatological care, and as such, it is important to evaluate the underlying skin disease classification algorithm on diverse skin tones to address potential accuracy disparities given the context of skin tone and race in the United States healthcare system and computer vision applications. However the algorithm used in Jain et al 2021 was trained on only 46 images of FST VI skin and 510 images of FST V, which was 0.3\% and 3.2\% of the entire training set, respectively~\cite{liu_deep_2020}. The lack of images of dark skin types in this dataset means this model may be prone to a higher level of unexpected algorithmic errors on future images of dark skin types~\cite{groh_evaluating_2021}. 

\section{Methods for Fitzpatrick Skin Type Annotations}

The Fitzpatrick labeling system is a six-point clinical scale used by dermatologists for classifying skin types based on photo-reactivity of skin and was originally intended to be used for photochemotherapy~\cite{fitzpatrick1988validity}. See Table~\ref{table:fitzscale} in Appendix for a copy of the original description of the Fitzpatrick Scale. We note that the original scale does not include nuanced skin tones or color beyond white, brown, and black. While the Fitzpatrick scale is highly correlated with an individual's melanin index (measured by narrow-band spectrophotometric
devices), the Fitzpatrick scale is a subjective measure~\cite{khalid2017utility}. In clinical practice, clinical FSTs are visually assessed by dermatologists based on the colors of a patient's skin, hair, and eyes and their history of sunburns~\cite{noauthor_fitzpatrick_2015}. In a study comparing self-reports to a single dermatologist's clinical FST determination, the dermatologist's assessment was found to be significantly more reliable than individuals' self reports~\cite{eilers2013accuracy}. Recently, researchers have used the estimated FST to annotate images and evaluate algorithmic fairness of AI models across apparent skin tones~\cite{buolamwini2018gender,liu_deep_2020,jain_development_2021,daneshjou_lack_2021, groh_evaluating_2021}. While the original Fitzpatrick scale was not designed to categorize skin color, it is often used as such in clinical practice~\cite{ware_racial_nodate} and it serves as a starting point (albeit imperfect given the coarseness of categories for skin color along sepia tones) for algorithmic audits~\cite{buolamwini2018gender}.

\subsection{Expert Labels from Board-Certified Dermatologists}

We asked three board-certified dermatologists – experts with deep experience examining skin conditions and assessing patients' clinical FST – to annotate images with the estimated FST. Each expert provided independent estimated FST annotations for 320 images collected from dermatology textbooks and 160 images collected from online dermatology Atlases. We collected 1,380 estimated FST annotations from experts. Given the inherent subjectivity of this task, we present ranges of experts' annotation across these images: 3-5\% Type I, 28-31\% Type II, 29-30\% Type III, 14-15\% Type IV, 14-15\% Type V, and 4-9\% Type VI. Likewise, the distribution across the the 160 images is 4-20\% unknown, 0-3\% Type I, 17-28\% Type II, 26-34\% Type III, 20-24\% Type IV, 9-13\% Type 5, and 1-8\% Type 6. 

The experts noted that estimated FSTs will not necessarily match in-person assessments because clinical FST relies on not just skin color but eye color, hair type and color, and history of sunburns. Moreover, clinical FST based on an in-person assessment considers an individual's entire body across varying lighting conditions while estimated FSTs based on a single image are restricted to a limited view of the body under a single lighting condition. As such, image-based estimated FST assessments will be have less information and be fundamentally more noisy with less inter-annotator agreement than clinical, in-person assessments. For example, clinical images of dermatological conditions differ in what part of the body is photographed, how the photograph is framed (from the camera's angle and zoom level to the patient's pose), how the lighting illuminates the image, and how the skin disease has transformed the patient's skin. We discuss further limitations of estimated FST annotations in the Limitations section.

\subsection{Algorithmic Labels from Individual Typology Angle}

Following computer vision papers using ITA-FST for algorithmic audits~\cite{kinyanjui_estimating_2019, krishnapriya2022analysis}, we compute ITA-FST annotations for each image. ITA was designed to classify skin color in a Caucasian population based on healthy skin in an image~\cite{chardon1991skin}. While ITA was not designed for all people, research shows ITA-FST correlates with both Melanin Index and clinical FST~\cite{noauthor_fitzpatrick_2015, del2013variations}. However, ITA-FST and clinical FST are designed to measure constitutive pigmentation and sun-reactivity, respectively, and recent research suggests they are poor proxies of one another~\cite{osto2022individual}. In order to calculate ITA-FST more precisely, researchers developed YCbCr masks to mask pixels outside a range of pre-specified colors~\cite{kolkur_human_2017} to reduce the noise of ITA-FST estimates. YCbCr masks are imperfect and often mask healthy skin or fail to mask non-skin parts of an image in the range of skin colors, but without YCbCr masks ITA-FST estimates are even more varied because very light or very dark backgrounds can influence the estimate. For example, YCbCr often fails to mask white underwear of dark skin people leading to the ITA-FST algorithm making errors in estimating skin tone that a reasonable human would not make.

We calculate ITA using the default D65 illuminant over the healthy skin pixels identified by YCbCr masks, and we convert the scores to estimated FSTs that minimize discrepancy between algorithmic labels and the experts' labels on the 320 textbook images and 140 images from dermatology atlases following procedures described by Groh et al 2021 and Krishnapriya et al 2022 ~\cite{groh_evaluating_2021, krishnapriya2022analysis}. See Algorithm~\ref{alg:thresholds} in Appendix for details on transforming ITA scores to FSTs.

\subsection{Dynamic Consensus Protocol for Crowd Labels}

In order to crowdsource estimated FST annotations for images, we collaborated with Scale AI and Centaur Labs, two companies that specialize in labeling large image datasets via dynamic consensus protocols applied to crowdworkers' annotations. In this section, we identify five key components of dynamic consensus protocols based the processes at Centaur Labs.

Dynamic consensus refers to the process of transforming multiple annotations from independent sources at different times on a single image into a consensus annotation. A dynamic consensus differs from a standard consensus metric like a mean, median, or mode because a dynamic consensus pre-specifies a \textbf{consensus threshold}, which must be met before annotations are transformed into accepted responses. For example, the annotations produced by Centaur Labs included a consensus threshold defined as either (a) a single category (across the 6 FSTs and a category for not applicable) has 3 more annotations than any other category or (b) the majority label if a consensus has not been reached after 20 annotations. 

\textbf{Qualified annotations} are defined as annotations by individuals who have passed a task specific quality control procedure. In contrast, disqualified annotations are annotations by individuals who have failed the task specific quality control. The third category, non-qualified annotations, are annotations by individuals who have not yet been assessed by a task specific quality control procedure. In general, quality control is determined by the proportion of an individual's annotations that correspond to a set of expert annotations. Given the subjectivity of estimated FST, we use both an expert's annotations to compare against 320 annotations collected from both Scale AI and Centaur Labs and the dynamic crowd consensus annotations on the rest of the images as measures on which to evaluate annotation quality. For both Scale AI and Centaur Labs, we seed the dynamic crowd consensus protocol with expert annotations to avoid crowd prejudice equilibria that can arise in cold-start annotation tasks~\cite{della_penna_crowd_2012}. In the dynamic crowd consensus protocol devised with Centaur Labs, we included a qualified minimum agreement of 40\% and qualified minimum and maximum annotations at 25 and 50, which means an individual is qualified only after attaining 40\% agreement on 25 images and then an individual only remains qualified as long as her agreement remains above 40\% for the 50 most recently annotated images. We selected the 40\% minimum agreement threshold with Centaur Labs for two reasons: first, it is significantly above random guessing, which would be 16.7\%, and second, we had previously found that 48\% of consensus annotations by Scale AI matched expert 1's annotation exactly, so we rounded down to the nearest multiple of ten. The qualified minimum and maximum annotation levels were suggested by Centaur Labs based on past performance of their crowdworkers on other similar datasets. We did not include a dynamic quality control procedure for (dis)qualifying annotations with Scale AI.

For the 320 textbook images, Scale AI provided 156,566 annotations (ranging from 378 to 1094 annotations per image) and Centaur Labs provided 7,999 qualified annotations (ranging from 2 to 93 qualified annotations per image). For an additional 16,577 images from the Fitzpatrick 17k dataset, Scale AI provided 62,710 annotations (with an interquartile range of 4 to 4 annotations per image) and Centaur Labs provided 265,279 qualified annotations (with an interquartile range of 9 to 20 qualified annotations per image)~\cite{groh_evaluating_2021}. In total, we collected 492,554 estimated FST annotations from crowd workers. 

In addition to estimated FST annotations, we collected \textbf{agreement metrics} for measuring the agreement and difficulty of annotating images with estimated FSTs. These agreement metrics are weighted by each individual annotator's agreement with the expert annotations and defined for each image as follows: agreement is the weighted, qualified annotations with the consensus label divided by the weighted, qualified annotations; difficulty is the weighted, qualified annotations without the consensus label divided by the weighted, qualified annotations. In algebraic notation, agreement and difficulty can be written as $A = \frac{Q_c}{Q}$ and difficulty as $D = \frac{Q\not\in Q_c}{Q}$ where $Q_c$ is the weighted number of qualified annotations with the consensus label, $Q$ is the weighted number of qualified annotations, and $Q\not\in Q_c$ is the weighted number of qualified annotations that are not the consensus label. 

Another criteria for assessing and improving the reliability of an images' annotations is the incorporation of \textbf{failure reports}~\cite{cabrera_discovering_2021}. Failure reports are comments on flagged images by annotators indicating that an image is either incorrectly labeled or inappropriate or irrelevant. Failure reports allow crowdsourced workers to transcend the 7 multiple choice labels (the FST scale and the not applicable option) to provide text-based feedback on the image. In the annotations by Centaur Labs, we stop labeling any image which was flagged as inappropriate or irrelevant once or flagged as incorrect twice. Across, the 320 textbook images, we received 20 failure reports on 17 images. We discuss the details of these failure reports in Section 4.2.

The final criteria for crowdsourcing is \textbf{expert review}, which is particularly useful for focusing the efforts by experts on the edge cases with high disagreement among crowd annotators. Expert review consists of experts reviewing flagged images without seeing the distribution of labels to adjudicate the annotation. We discuss the results of expert review in Section 4.3. 

\section{Results Comparing Annotations on 320 Textbook Images}

We asked three board-certified dermatologists to annotate 320 images with FSTs, and we find that the annotations of any two experts match exactly on 50-55\% of images and match within one unit on 92-94\% of images. Figure~\ref{fig:cm-e1-e2} presents a confusion matrix comparing the annotations of the first two experts.

In comparison to the two experts' labels, the algorithmically generated annotations for the 320 images are much less similar. The ITA-FST algorithm produces Fitzpatrick labels identical to expert 1, 2, and 3 in 27\%, 31\%, and 40\% of images, respectively, and is off by no more than a single unit (i.e., FST I vs FST II) in 70\%, 69\%, and 76\% of images, respectively. See Figure~\ref{fig:cf_matrices} and Figure~\ref{fig:cf_matrices_3} in the Appendix for confusion matrices examining annotation discrepancies between experts 1 and 2 and the Scale AI, Centaur Labs, and ITA-FST algorithm.

The inter-rater reliability between the two experts' and the crowds' annotations is much more similar across the 320 images. The labels produced by Scale AI and Centaur Labs match expert 1 exactly in 48\% and 40\% of images, match expert 2 exactly in 50\% and 38\% of images, and match expert 3 exactly in 58\% and 43\% of images, all respectively. Likewise, the annotations produced by Scale AI and Centaur Labs are off by no more than a single unit from expert 1's annotations in 94\% and 87\% of images, expert 2's annotations in 91\% and 79\% of images, and expert 3's annotations in 93\% and 88\% of images. 

\begin{center}
\begin{figure*}[h]
\def\myConfMat{{
{0,  1,  7,  1,  1,  0,  0},  
{0,  5,  5,  0,  0,  0,  0},  
{0,  9, 50, 36,  5,  0,  0},  
{0,  1, 24, 50, 17,  5,  0},  
{0,  0,  2,  7, 20, 17,  1},  
{0,  0,  0,  0,  2, 25, 17},  
{0,  0,  0,  0,  0,  1, 12},  
}}

\def\classNames{{"NA","1","2","3","4","5","6"}} 

\def\numClasses{7} 

\def\myScale{1} 
\begin{tikzpicture}[
    scale = \myScale,
    ]

\tikzset{vertical label/.style={rotate=90,anchor=east}}   
\tikzset{diagonal label/.style={rotate=45,anchor=north east}}

\foreach \y in {1,...,\numClasses} 
{
    \node [anchor=east] at (0.4,-\y) {\pgfmathparse{\classNames[\y-1]}\pgfmathresult}; 
    
    \foreach \x in {1,...,\numClasses}  
    {
    \def\totSamples{0}
    \foreach \ll in {1,...,\numClasses}
    {
        \pgfmathparse{\myConfMat[\ll-1][\x-1]}   
        \xdef\totSamples{\totSamples+\pgfmathresult} 
    }
    \pgfmathparse{\totSamples} \xdef\totSamples{\pgfmathresult}  


    
    \begin{scope}[shift={(\x,-\y)}]
        \def\mVal{\myConfMat[\y-1][\x-1]} 
        \pgfmathtruncatemacro{\r}{\mVal}   %
        \pgfmathtruncatemacro{\p}{round(\r/(0.0001+\totSamples)*100)}
        \coordinate (C) at (0,0);
        \ifthenelse{\p<50}{\def\txtcol{black}}{\def\txtcol{white}} 
        \node[
            draw,                 
            text=\txtcol,         
            align=center,         
            fill=black!\p,        
            minimum size=\myScale*10mm,    
            inner sep=0,          
            ] (C) {\r\\\p\%};     
        \ifthenelse{\y=\numClasses}{
        \node [] at ($(C)-(0,0.75)$) 
        {\pgfmathparse{\classNames[\x-1]}\pgfmathresult};}{}
    \end{scope}
    }
}
\coordinate (yaxis) at (-0.3,0.5-\numClasses/2);  
\coordinate (xaxis) at (0.5+\numClasses/2, -\numClasses-1.25); 
\node [vertical label] at (yaxis) {Expert 1};
\node []               at (xaxis) {Expert 2};

\end{tikzpicture}
\caption{Confusion matrix comparing two board-certified dermatologists' Fitzpatrick skin type annotations on 320 images from dermatology textbooks.
}
\label{fig:cm-e1-e2}
\end{figure*}
\end{center}

\subsection{Quantitative Assessment of Inter-Rater Reliability on 320 images}

In light of the subjectivity of estimated FST annotations, we evaluate annotation performance by comparing inter-rater reliability between pairs of experts with the inter-rater reliability between experts and each non-expert annotation method. Specifically, we measure inter-rater reliability using the Pearson Correlation Coefficient ($\rho$) between two annotation methods, and we evaluate the statistical significance following the Fisher Z transformation for comparing independent correlations~\cite{fisher1915frequency}. We describe the pseudocode for comparing $\rho_{E_i,E_j}$ with $\rho_{E_{i,Method}}$ in Algorithm~\ref{alg:paircompare} in the Appendix where $E_i$ and $E_j$ refer to one of the three experts and $E_{Method}$ refers to one of the non-expert annotation methods. When calculating the $\rho_{X,Y}$ between two annotation methods X and Y, we drop annotations that either method marks as not applicable.

\begin{figure*}[h]
    \centering
    \includegraphics[width=0.45\textwidth]{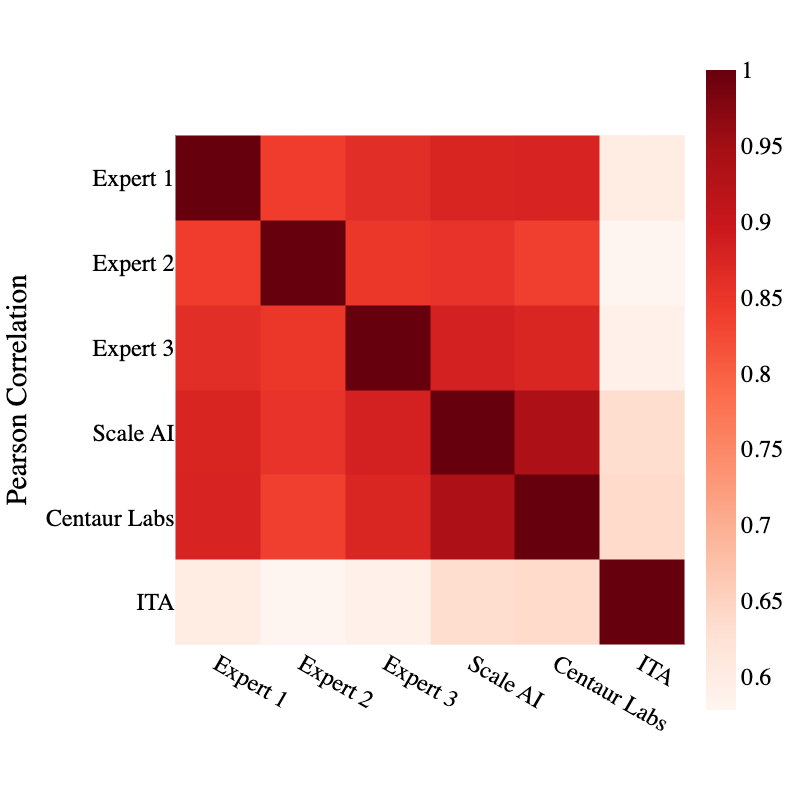}
    \qquad
    \includegraphics[width=0.45\textwidth]{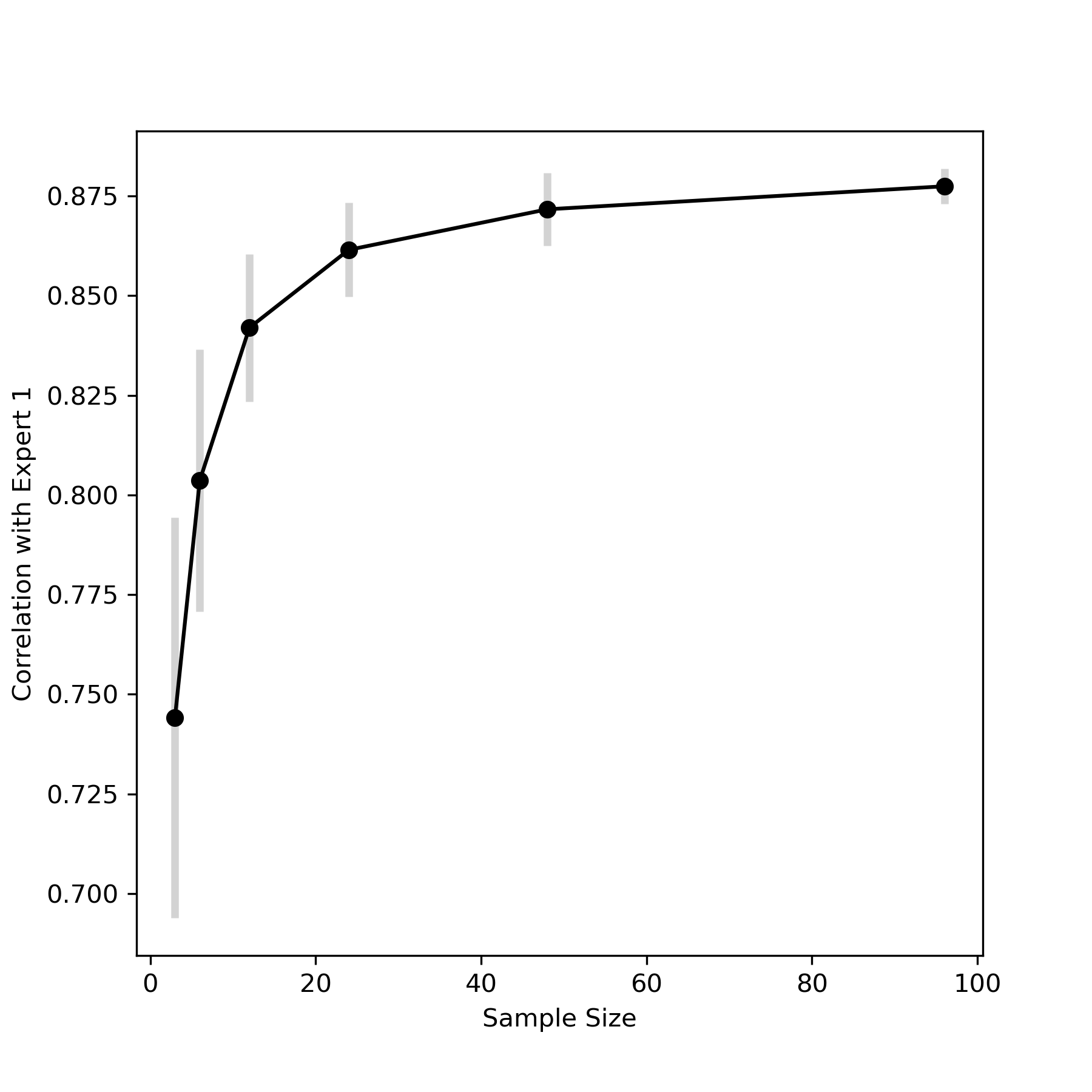}
    \caption{\textbf{Left}: Heatmaps showing inter-rater reliability as measured by Pearson's Correlation Coefficient. These heatmaps include 296 images and exclude the 24 images rated by any expert or crowdsourcing method as ``Not Applicable.'' \textbf{Right}: Inter-rater reliability by crowd size based on 25 random bootstrapped samples from the Scale AI annotations. The y-axis presents the correlation between expert 1's annotations and the crowd's mean FST annotation. The x-axis presents the number of annotations per image. The gray bars represent the 95\% confidence interval. As the number of annotations increases the confidence interval decreases and the Pearson Correlation Coefficient ($\rho$) approaches 0.88.}
    \label{fig:pearson}
\end{figure*}

We find the inter-rater reliability of the ITA-FST algorithm is significantly lower than the inter-rater reliability of experts. The correlation between the first two experts' annotations is $\rho_{E_1,E_2}=.84$ ($E_1$ and $E_2$ refer to expert 1 and 2, respectively) whereas the correlation between the ITA-FST algorithm and any of the experts is $\rho_{ITA,E_1}=.57$, $\rho_{ITA,E_2}=.52$, and $\rho_{ITA,E_3}=.55$. The differences between any pair of experts $\rho_{E_i,E_j}$ and $\rho_{E_1,ITA}$, $\rho_{E_2,ITA}$, and $\rho_{E_3,ITA}$ are statistically significant ($p<0.00000001$). We present the correlations and the p-value of the comparisons of correlations in Table~\ref{table:rezzie}. We also present a heatmap of inter-rater reliability as measured by $\rho$ in Figure~\ref{fig:pearson}. 

In contrast to the low inter-rater reliability between experts and the algorithm, we find the inter-rater reliability of expert and crowdsourced annotations to be comparable. Notably, the crowdsourced annotations are slightly more correlated with experts' annotations in five of six comparisons – $\rho_{E_1,S}=.88$, $\rho_{E_1,C}=.88$, $\rho_{E_2,S}=.86$, $\rho_{E_2,C}=.83$, $\rho_{E_3,S}=.87$, $\rho_{E_3,C}=.87$ (S and C refer to Scale AI and Centaur Labs, respectively) – than experts' annotations are correlated with each other ($\rho_{E_1,E_2}=.84$, $\rho_{E_2,E_3}=.85$, and $\rho_{E_1,E_3}=.86$). We do not find statistically significant differences between the experts' correlation with each other and either expert's correlation with any of the crowdsourced methods. 

\begin{center}
\begin{table}[h]
\begin{tabular}{ l l l l l l l}
  \toprule
 Method & $E_1$ ($\rho$) & $E_1$ p-value &  $E_2$ ($\rho$) & $E_2$ p-value &  $E_3$ ($\rho$) & $E_3$ p-value \\ 
 \toprule
 $E_1$ &  & & 0.84 & 0.73 & 0.86 & 0.66 \\ 
 $E_2$ & 0.84  & 0.44 & & & 0.85 & 0.66 \\ 
 $E_3$ & 0.86  & 0.44 & 0.85 & 0.73  \\ 
 ITA-FST & 0.57 & <0.001 & 0.52 & <0.001 & 0.55 & <0.001 \\
 Scale AI & 0.88 & 0.08 & 0.86 & 0.43 & 0.88 & 0.08  \\
 Centaur Labs & 0.88 & 0.08 & 0.83 &  0.50 & 0.87 & 0.32 \\
 \bottomrule
 \vspace{.05cm}
\end{tabular}
\caption{Inter-rater reliability based on Fisher Z transformations of Pearson Correlation Coefficients ($\rho$). The $E_x$ ($\rho$) columns display the correlation between the method in the row and the method in the column. The p-value columns show the minimum p-value based on Algorithm~\ref{alg:paircompare} in the Appendix applied to all pairwise correlations of experts; as an example, the cell in the $E_1$ p-value column and ITA-FST row presents the minimum p-value comparing (a) $\rho_{E_1,ITA}$ and $\rho_{E_1,E_2}$,  (b) $\rho_{E_1,ITA}$ and $\rho_{E_1,E_3}$, and (c) $\rho_{E_1,ITA}$ and $\rho_{E_2,E_3}$}.
\label{table:rezzie}
\end{table}
\end{center}

In addition to examining the inter-rater reliability across methods, we examine how inter-rater reliability changes depending on the number of non-qualified annotations. Instead of assessing FSTs based on a dynamic consensus procedure, we compare expert 1's annotations with the crowd mean of 25 random draws from the Scale AI annotations (which were non-qualified meaning that crowdworkers were not filtered by a task specific quality control procedure) in samples of the following sizes: 3, 6, 12, 24, 48, and 96 annotations. We find a logarithmic relationship between $\rho_{S,E_1}$ and sample size that plateaus with $\rho_{S,E_1}$ approaching 0.88; see Figure~\ref{fig:pearson} for a visualization of this relationship. For example, an increase from 3 to 12 annotations per image is associated with a 10 percentage point increase in $\rho_{S,E_1}$; the mean $\rho_{S,E_1}$ is 0.74 with a standard deviation of 0.026 when evaluating across 3 annotations per image and 0.84 with a standard deviation of 0.01 when evaluating across 12 annotations per image. A further increase from 12 to 24 annotations per image is associated with another 2 percentage points increase in $\rho_{S,E_1}$. We also find a similar relationship when comparing the varying size of the crowd with expert 2 and 3.

\subsection{Qualitative Assessment of Inter-Rater Reliability}

\begin{figure*}[h]
    \centering
    \includegraphics[width=0.45\textwidth]{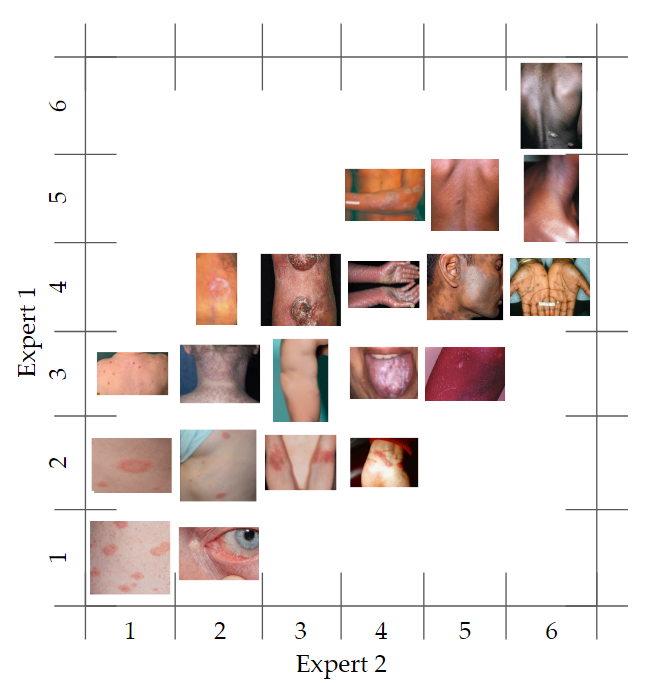}
    \includegraphics[width=0.45\textwidth]{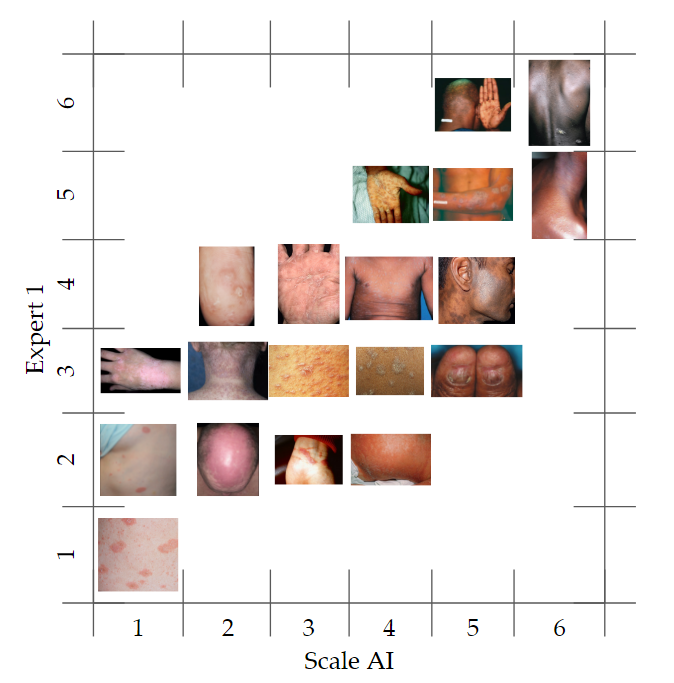}
    \includegraphics[width=0.45\textwidth]{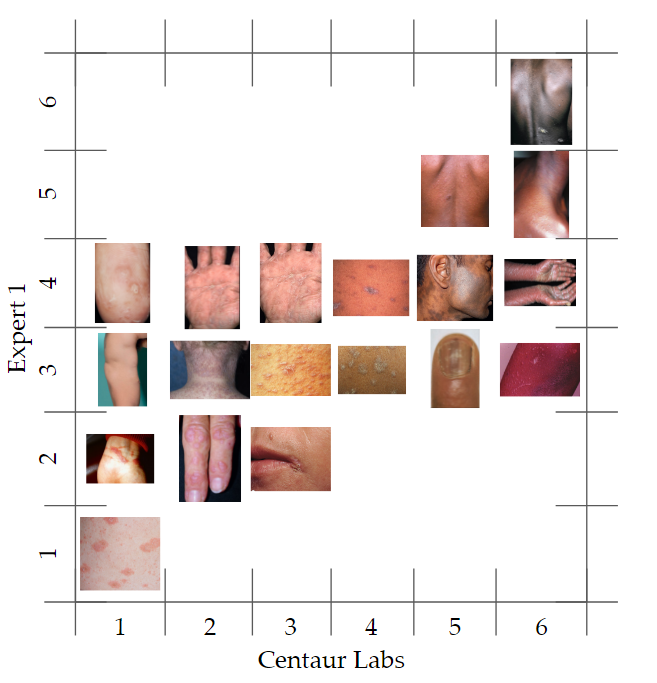}
    \includegraphics[width=0.45\textwidth]{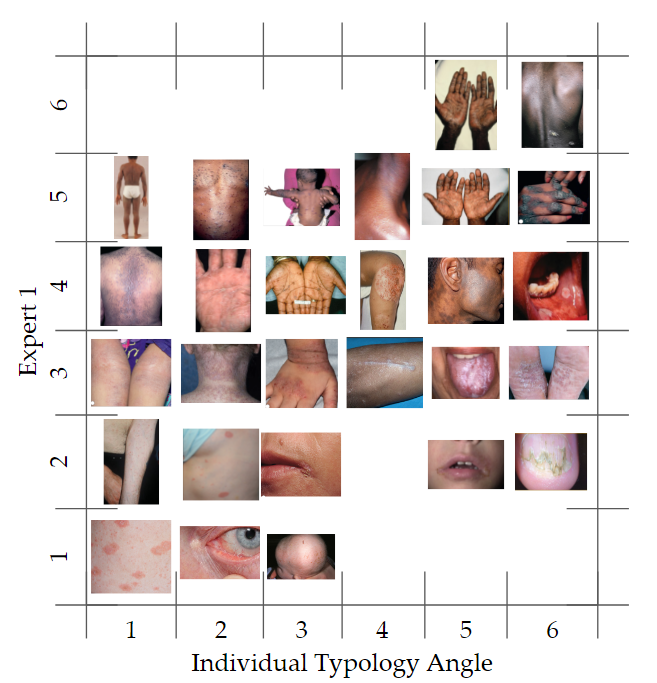}
    \caption{Textbook images of skin conditions plotted according to Expert 1's annotations (on the Y-axis) and 4 other methods (Expert 2, ITA-FST algorithm Scale AI, and Centaur Labs on the X-axis).}
    \label{fig:f2}
\end{figure*}

We examine inter-rater reliability qualitatively by illustrating similarities and differences in annotations across methods and examining images flagged by failure reports. In Figure~\ref{fig:f2}, we present \textbf{qualitative confusion matrices} that showcase how different annotation methods lead to different annotations. These qualitative confusion matrices are intended to contextualize and illustrate similarities and discrepancies in subjective annotations and build upon the finding that alternative representation of confusion matrices can improve non-expert understanding of performance~\cite{shen2020designing}.

Across the 320 textbook images, annotators flagged 17 images as inappropriate or incorrect. Three of these flagged images were originally marked by expert 1 as "Not Applicable." Unlike most images, all three of these images contain multiple photographs under multiple lighting conditions, expert 2 provided a different annotation than expert 1, and the Centaur Labs and Scale AI crowd labels are discordant. Another two of these flagged images are marked as confusing and neither the expert annotations or the crowd annotations agree with one another. The final 12 of the flagged images contain messages that the annotator is confident that the expert's label is wrong; in 5 of these 12 images, expert 2 and both crowd consensuses agree that expert 1's annotation is one unit off, in 6 of these 12 images, expert 2 agrees with expert 1 while both crowd consensuses disagree with the experts, and in 1 of these 12 images, there is disagreement across experts and crowd consensuses. These results suggest failure reports are generally useful in identifying images that are likely to be problematic and extremely subjective for one reason or another.

\subsection{Scaling Annotations on the Fitzpatrick 17k}

For resource constrained developers of large-scale image datasets, it is orders of magnitude less resource intensive to annotate images with an algorithm or crowdsourcing than with board-certified dermatologists~\cite{sachdeva2020gender, tucker2018ethical, orting2019survey}. Given the lower inter-rater reliability of the ITA-FST algorithm, we limit our analysis of scaling annotations on the full Fitzpatrick 17k dataset~\cite{groh_evaluating_2021} to crowdsourcing methods. 85\% of consensus FST annotations by Scale AI and Centaur Labs are within one unit of each other. In Figure~\ref{fig:cm-c-s-fitz17}, we present a confusion matrix, which reveals that large discrepancies in annotations between sources are rare. An expert review of all applicable annotation discrepancies that are off by more than one unit would involve examining 9\% (1,365 of the 13,865 images) of the Fitzpatrick 17k dataset. Error reports by annotators from Centaur Labs indicate that the consensus annotation for 166 images are incorrectly labeled and 21 images are inappropriate or irrelevant for the task. 

\begin{center}
\begin{figure*}[h]
\def\myConfMat{{
{291,   75,   87,   33,   34,   31,   14},  
{139, 2461,  291,   42,    9,    4,    1},  
{275, 2400, 1559,  491,   67,   12,    4},  
{207,  492,  880, 1288,  394,   43,    4},  
{110,   97,  302,  949,  968,  327,   28},  
{43,   18,   52,  122,  470,  667,  161},  
{8,   18,    9,    9,   32,  209,  350},  
}}

\def\classNames{{"NA","1","2","3","4","5","6"}} 

\def\numClasses{7} 

\def\myScale{1} 
\begin{tikzpicture}[
    scale = \myScale,
    ]

\tikzset{vertical label/.style={rotate=90,anchor=east}}   
\tikzset{diagonal label/.style={rotate=45,anchor=north east}}

\foreach \y in {1,...,\numClasses} 
{
    \node [anchor=east] at (0.4,-\y) {\pgfmathparse{\classNames[\y-1]}\pgfmathresult}; 
    
    \foreach \x in {1,...,\numClasses}  
    {
    \def\totSamples{0}
    \foreach \ll in {1,...,\numClasses}
    {
        \pgfmathparse{\myConfMat[\ll-1][\x-1]}   
        \xdef\totSamples{\totSamples+\pgfmathresult} 
    }
    \pgfmathparse{\totSamples} \xdef\totSamples{\pgfmathresult}  
    
    \begin{scope}[shift={(\x,-\y)}]
        \def\mVal{\myConfMat[\y-1][\x-1]} 
        \pgfmathtruncatemacro{\r}{\mVal}   %
\pgfmathtruncatemacro{\p}{round(\r/(0.0001+\totSamples)*100)}
        \coordinate (C) at (0,0);
        \ifthenelse{\p<50}{\def\txtcol{black}}{\def\txtcol{white}} 
        \node[
            draw,                 
            text=\txtcol,         
            align=center,         
            fill=black!\p,        
            minimum size=\myScale*10mm,    
            inner sep=0,          
            ] (C) {\r\\\p\%};     
        \ifthenelse{\y=\numClasses}{
        \node [] at ($(C)-(0,0.75)$) 
        {\pgfmathparse{\classNames[\x-1]}\pgfmathresult};}{}
    \end{scope}
    }
}
\coordinate (yaxis) at (-0.3,0.5-\numClasses/2);  
\coordinate (xaxis) at (0.5+\numClasses/2, -\numClasses-1.25); 
\node [vertical label] at (yaxis) {Scale AI};
\node []               at (xaxis) {Centaur Labs};

\end{tikzpicture}
\caption{Confusion matrix comparing two crowdsourcing methods for annotating the 16,577 images in the Fitzpatrick 17k dataset}
\label{fig:cm-c-s-fitz17}
\end{figure*}
\end{center}

\subsection{Expert Review of Scaled Annotations}

As a final step in evaluating dynamic consensus protocols, we collect labels from 3 board-certified dermatologists on 140 images randomly selected from the 16,577 images in the Fitzpatrick 17k dataset. We stratified this random selection on two features: (1) Scale AI's estimated FST annotations and (2) a binary variable for discrepancy between Scale AI's and Centaur Labs' annotations of more than 1 estimated FST annotation. As a result, there are 20 images with each Scale AI estimated FST type and 20 images annotated by Scale AI as not applicable. In addition, 70 of these images have been annotated by Scale AI and Centaur Labs within 1 estimated FST of each other and the other 70 images have been annotated with estimated FST that differ by more than 1.

For the 70 images with similar annotations, the correlation between experts ranges from 83\% to 87\% and the crowds correlation with experts ranges from 86\% to 89\%. We do not see any statistically significant difference between experts and crowds.

However, for the 70 images with greater than 1 unit discrepancies across the two crowd methods, we do find significant differences between inter-annotator reliability across experts and the crowd. The correlation between experts ranges from 59\% to 66\% and the crowds' correlation with experts ranges from 32\% to 63\%. We examine the inter-rater reliability between Scale AI and Centaur Labs and experts by conducting 12 tests of statistical significance to cover all possible comparison permutations. We find that 3 of 6 comparisons of inter-rater reliability between Scale AI and experts show Scale AI's annotations are less correlated and the p-value is less than the standard 5\% threshold for statistical significance. Likewise, we find that 1 of 6 comparisons of inter-rater reliability between Centaur Labs and experts show Centaur Labs' annotations are less correlated and the p-value is less than the standard 5\% threshold for statistical significance. In Table~\ref{table:rezzie140}, we present the inter-rater reliability Pearson correlation coefficients and lowest p-values for tests of statistical significance. This table also includes an examination of estimated ITA-FST on these 70 images, and we find estimated the correlation between ITA-FST and experts' annotations approaches 0 for this selection of images for expert review.

\begin{center}
\begin{table}[h]
\begin{tabular}{ l l l l l l l}
  \toprule
 Method & $E_1$ ($\rho$) & $E_1$ p-value &  $E_2$ ($\rho$) & $E_2$ p-value &  $E_3$ ($\rho$) & $E_3$ p-value \\ 
 \toprule
 $E_1$ &  & & 0.59 & 0.29 & 0.66 & 0.29 \\ 
 $E_2$ & 0.59 & 0.29 & & & 0.66 & 0.58 \\ 
 $E_3$ & 0.66  & 0.29 & 0.66 & 0.58  \\ 
 ITA-FST & 0.05 & <0.001 & -0.06 & <0.001 & 0.08 & <0.001\\
 Scale AI & 0.50 & 0.04 & 0.32 & <0.001 & 0.57 & 0.19  \\
 Centaur Labs & 0.63 & 0.54 & 0.47 &  0.02 & 0.53 & 0.09 \\
 \bottomrule
 \vspace{.05cm}
\end{tabular}
\caption{Analysis of subset of 70 images with high disagreement showing the inter-rater reliability based on Fisher Z transformations of Pearson Correlation Coefficients ($\rho$). The $E_x$ ($\rho$) columns display the correlation between the method in the row and the method in the column. The p-value columns show the minimum p-value based on Algorithm~\ref{alg:paircompare} in the Appendix applied to all pairwise correlations of experts; as an example, the cell in the $E_1$ p-value column and ITA-FST row presents the minimum p-value comparing (a) $\rho_{E_1,ITA}$ and $\rho_{E_1,E_2}$,  (b) $\rho_{E_1,ITA}$ and $\rho_{E_1,E_3}$, and (c) $\rho_{E_1,ITA}$ and $\rho_{E_2,E_3}$}.
\label{table:rezzie140}
\end{table}
\end{center}

The comparison of inter-annotator agreement on the images selected for expert review reveals important nuances that researchers should keep in mind when annotating future datasets. While the inter-rater reliability on estimated FST is just as high between experts as it is between experts and the crowd consensus for most images, the annotations by crowds on images with low agreement may be less reliable than experts' annotations. By incorporating \textbf{expert review} into a subset of crowd annotations with low agreement, a dynamic consensus protocol can adjudicate edge cases such that adjudication leads to a higher likelihood of agreement with other experts.

This particular expert review of 140 images highlights edge cases where experts tend to agree with each other more often than they agree with dynamic consensus labels from crowdworkers. However, it is important to note that inter-rater reliability across experts on the 70 images randomly drawn from the 9\% of images with two discordant crowd ratings ranges from 59\% to 66\% whereas inter-rater reliablity on the other 70 images (randomly drawn from the 91\% of images with two concordant crowd ratings) ranges from 83\% to 87\%. On these 70 images with discordant crowd annotations, experts agree with each other significantly less than they do on the overwhelming majority of images. This lower rate of expert agreement and significantly lower rate of crowd worker and expert agreement demonstrates the subjectivity of estimating FST of an individual in an image can vary considerably across images.

\section{Discussion}

How well does the ITA-FST algorithm and various crowdsourcing methods compare to board-certified dermatologists in annotating images with estimated FSTs? Our results reveal that the inter-rater reliability between three board-certified dermatologists (as measured by $\rho_{E_x,E_y}$) is comparable to the inter-rater reliability between each board-certified dermatologist and each of the two crowdsourcing methods (as measured by $\rho_{E,Crowd}$). However, inter-rater reliability of the ITA-FST algorithm (as measured by $\rho_{E,ITA}$) is significantly lower than the inter-rater reliability between any two experts. 

Estimated FST annotations on images are highly subjective. We find that three experts agree with each other exactly on estimated FST in only 50-55\% of images (although they agree with each other within a one unit difference in 92-94\% of images). Rather than treat this subjectivity as a bias, we treat subjectivity on a per annotation basis as a measure of signal and noise. We find the differences between the three experts' annotations are not significantly larger than the differences between the experts and either crowdsourced method. In other words, expert annotations generally have the same amount of signal and noise as crowd annotations. This general finding comes with a caveat: there are identifiable edge cases where experts' annotations demonstrated significantly higher inter-rater reliability than crowdsourced annotations. Nonetheless, our results suggest that crowdsourcing methods (but not the ITA-FST algorithm) can be reliable for annotating large scale dermatology image datasets with skin type annotations especially when expert review is included. 

This is particularly important for increasing transparency in machine learning for dermatology because skin type annotations are one of the items on the CLEAR Derm checklist for the evaluation of image-based AI algorithms~\cite{daneshjou_checklist_2021} and an important consideration for evaluating medical AI devices for FDA approvals~\cite{wu2021medical}. Transparency on skin tone information can be useful for evaluating both the distribution (and potential under-representation) of various skin tones in image datasets and how AI algorithms in dermatology perform across different skin tones, which is then useful as evidence for holding the fields of computer vision and dermatology accountable for addressing the unwanted biases.

While crowdsourced annotations are comparable with experts' annotations in aggregate, there are many examples where experts agree with each other yet the crowd differs. One approach for reducing crowdsourcing disagreement with experts is to include more annotations per image, which we find is effective for reducing errors from crowd sizes of 3 to 12 but less effective for reducing errors from larger crowd sizes. A second approach is to integrate expert review into crowdsourcing. In particular, expert review examines edge cases that are flagged based on failure reports, agreement metrics (e.g. low agreement scores, high difficulty scores), and random samples for review. For the overwhelming majority of images, experts and the crowd have similar inter-rater reliability, but for the edge cases, expert review can offer additional reliability because inter-rater reliability of experts on edge cases can be higher than inter-rater reliability of crowds on edge cases. 

The comparison between methods for annotating subjective labels provides a replicable methodology for answering when an algorithm or crowdsourced methodology can reliably be used in lieu of experts for annotating data. The goal of this kind of data annotation is to increase transparency in dataset biases to motivate greater accountability in sociotechnical decision-making systems. However, this kind of transparency comes at a cost. Human labor by experts or crowd annotators requires time and energy and should be compensated appropriately whereas the resources needed to compute ITA-FST scores are neglible. The low agreement between ITA-FST and experts is best to avoid because it may leave analyses prone to data cascade errors~\cite{sambasivan2021everyone}. On the other hand, the relatively high agreement between experts and the crowd (and the opportunity to augment crowd labels with expert review) makes crowd annotations of estimated FST on images more attractive than expensive experts. We note that the crowd labels here come from Scale AI and Centaur labs, which represent very different ecosystems than the decentralized requester marketplace of Amazon Mechanical Turk (AMT)~\cite{martin2014being}. In particular, Scale AI and Centaur Labs work directly with individuals rather than through AMT, and as such, both these services avoid the ``root problem... of unfair requesters''~\cite{hara2018data} in the AMT marketplace and the problem of turkers' uncertainty about the fairness of a particular requester~\cite{irani2013turkopticon}. Moreover, the ability to submit error reports with Centaur Labs creates a tractable opportunity for communication between crowdworkers analyzing the images and data scientists analyzing the data. 

\section{Limitations}

We focused our comparisons on how three experts, one algorithm, and two crowdsourcing methods retrospectively annotate estimated FST across 320 images collected from dermatology textbooks and 140 images collected from online dermatology atlases. The 320 images showcase skin of all six skin types, but the distribution of skin types is not uniform across these images because dark skin types are underrepresented in dermatology textbooks~\cite{ebede_disparities_2006,alvarado_representation_2020,adelekun_skin_2020}. Based on experts' annotations, only 18-26\% of the 320 images show the two darkest skin types. 

In our evaluation, we consider the ITA-FST algorithm applied to images with YCbCr masks, and we find it exhibits higher variability than expert and crowd-based annotations. While the ITA-FST algorithm may not be a reliable method for annotating estimated FST, future algorithms applied to images (especially segmented portions of images) may be able to match the inter-rater reliability of experts. 

The lighting conditions are heterogeneous across these images, which makes assessing estimated FST more difficult than it would be in images with a single, consistent cross-polarized light source. Guidelines for photographing images of skin conditions on dark skin suggest images use indirect, natural light and a separate light for the hair and should avoid backgrounds with bright colors or patterns~\cite{lester2021clinical}. A recent perspective piece in the British Journal of Dermatology presents a series of images where the only difference is lighting source (cross-polarized light vs. white light) that reveals cross-polarized light reduces specular reflections and increases the contrast between healthy and unhealthy skin~\cite{oh2021standardized}. 

The variability of estimated FST annotations in images is much higher than in-person assessments because in-person assessments are not limited by lighting sources and enable a dermatologist to include an assessment of the patients' skin color, eye color, hair color, and history of sunburns. We leave the comparison of in-person FST annotation to image-based estimated FST annotations to future research.

The Fitzpatrick scale is a starting point but not a perfect method for annotating skin color~\cite{buolamwini2018gender,ware_racial_nodate}. The Fitzpatrick classification system was originally designed for classifying skin based on skin's reaction to the sun (burning vs tanning) and not skin color~\cite{fitzpatrick1988validity}. Moreover, the original Fitzpatrick classification labeled FST I-IV as white, FST V as brown, and FST VI as black, which contrasts with how researchers describe today's usage of FST as pale-white for I, white for II, beige for III, brown for IV, dark brown for V, and black for VI~\cite{roy2022does}. We re-created the original scale in Table~\ref{table:fitzscale} in the Appendix for quick reference. Annotating images with estimated FSTs helps to document the diversity of dermatology datasets and inspect algorithms for discrimination based on the color of one's skin, but estimated FSTs serve as a blunt proxy (biased towards lighter skin colors) that fail to capture the global diversity of skin colors~\cite{ware_racial_nodate}. In order to avoid singularly optimizing future AI algorithms on a biased proxy~\cite{mullainathan2021inequity}, future research and data collection should consider additional methods and metrics for annotating the diversity and complexity of skin color including factors such as self-reported versus observer reported skin tone~\cite{monk2015cost}, in-person or image based assessment, and the number of response categories~\cite{preston2000optimal}.

\section{Conclusion}

By annotating large datasets of dermatology images with FSTs, researchers can increase transparency and enable relatively straight-forward evaluations of algorithmic performance across skin types for AI models trained to classify skin conditions. While image-based FST annotations are subjective, we find the annotations of experts and crowds are highly comparable while the annotations produced by the ITA-FST algorithm are more variable. In light of the higher variability of annotations generated by the ITA-FST algorithm, we recommend that researchers do not augment their datasets of clinical dermatology images algorithmically and instead use a crowdsourcing or expert-based approach. 

We find some instances where the experts concur yet the crowd consensus disagrees. We recommend the most efficient and thorough approach to annotating images of skin conditions with FSTs is to combine experts and the crowd. Expert review can adjudicate both images flagged for error reports and images with low agreement or high difficulty scores. While we propose this approach for annotation of FSTs, our recommendation for hybrid dynamic consensus protocols with experts and crowds may extend to other domains in which annotations are similarly subjective for experts and non-experts alike.

\section*{Data and Code Availability}

The datasets and code generated and analyzed during the current study are available in our public Github repository, \url{https://github.com/mattgroh/fitzpatrick17k}.

\section*{Acknowledgments}

We thank Erik Duhaime and Kira Prentice at Centaur Labs and Aerin Kim at Scale AI for providing data annotation services for free, the many crowdworkers for annotating the images, Bruke Wossenseged for valuable research assistance, and Rosalind Picard, Ziv Epstein, and Luis Soenksen for thoughtful feedback.

\bibliographystyle{ACM-Reference-Format}
\bibliography{citations}

\section*{Appendix}

\begin{algorithm}
	\caption{Fisher Z transformation for comparing independent correlations}
	\begin{algorithmic}[1]
		\For {Each Expert}
		\For {Each Method}
		\State $z_{E_A,E_B} \gets \frac{1}{2}\ln{\frac{1+\rho_{E_A,E_B}}{1-\rho_{E_A,E_B}}}$
		\State $z_{Expert,Method} \gets \frac{1}{2}\ln{\frac{1+\rho_{Expert,Method}}{1-\rho_{Expert,Method}}}$
		\State $Z \gets \lvert \frac{z_{E_A,E_B} - z_{Expert,Method}}{\sqrt{2/(n-3)}}\rvert$ where n= \# of images
		\State Convert Z score to p-value
		\EndFor
		\EndFor
	\end{algorithmic}
	\label{alg:paircompare}
\end{algorithm} 

\begin{algorithm}
	\caption{Individual typology angle threshold adjustment}
	\begin{algorithmic}[1]
	    \State $T_{12} \gets Mean(ITA_1.Quantile(1), ITA_2.Quantile(3))$ 
	    \State $T_{23} \gets Mean(ITA_2.Quantile(1), ITA_3.Quantile(3))$
	    \State $T_{34} \gets Mean(ITA_3.Quantile(1), ITA_4.Quantile(3))$
	    \State $T_{45} \gets Mean(ITA_4.Quantile(1), ITA_5.Quantile(3))$
	    \State $T_{56} \gets Mean(ITA_5.Quantile(1), ITA_6.Quantile(3))$
	    \State $T \gets \{T_{12}, T_{23}, T_{34}, T_{45}, T_{56}\}$
		\ForAll {$t \in T$}
		    \State $Max\_Concordant \gets Sum(Annotation\_E_1 = Annotation\_E_2 = ITA(t))$
		    \State $I \gets \{-5, -4, -3, -2, -1, 0, 1, 2, 3, 4, 5\}$
    		\ForAll {$i \in I$}
        		\State $t_i \gets t + i$
        		\State $Concordant \gets Sum(Annotation\_E_1 = Annotation\_E_2 = ITA(t_i))$
        		\If {$Concordant > Max\_Concordant$}
        		    \State $Max\_Concordant \gets Concordant$
        		    \State $t \gets t_i$
        		\EndIf
    		\EndFor 
		\EndFor 
	\end{algorithmic}
	\label{alg:thresholds}
\end{algorithm} 

\begin{center}
\begin{table}[h]
\begin{tabular}{ l l l l l l l}
  \toprule
 Skin Type & Skin Color & Sunburn & Tan &  \\ 
 \toprule
 I & White & Yes & No \\
 II & White & Yes & Minimal \\
 III & White & Yes & Yes \\
 IV & White & No & Yes \\
 V & Brown & No & Yes \\
 VI & Black & No & Yes \\
 \bottomrule
 \vspace{.05cm}
\end{tabular}
\caption{The Fitzpatrick skin type scale from Fitzpatrick et al 1988~\cite{fitzpatrick1988validity}. The scale is intended for classifying sun-reactive skin types. Notably, the original scale does not include nuanced skin tones or color beyond white, brown, and black. In dermatology practice, the Fitzpatrick scale is commonly used to describe constitutive skin color~\cite{ware_racial_nodate}. Recent research published in the \textit{Medical Image Analysis} describes the Fitzpatrick skin types as pale-white, white, beige, brown, dark brown, and black~\cite{roy2022does}. We informed crowd annotators by presenting example images of each FST.}
\label{table:fitzscale}
\end{table}
\end{center}



\begin{figure}[p]

    \begin{multicols}{2}
        \def\myConfMat{{
        {3,  1,  6,  0,  0,  0,  0},  
        {0,  9,  1,  0,  0,  0,  0},  
        {0,  25, 69, 5,  1,  0,  0},  
        {1,  5, 48, 24, 16,  4,  0},  
        {0,  0,  1,  7, 17, 21,  1},  
        {0,  0,  0,  0,  2, 21, 21},  
        {0,  0,  0,  0,  0,  2, 11},  
        }}
        \def\classNames{{"NA","1","2","3","4","5","6"}} 
        \def\numClasses{7}
        \def\myScale{0.7}
        \begin{tikzpicture}[
            scale = \myScale,
            font={\footnotesize}, 
            ]
        \tikzset{vertical label/.style={rotate=90,anchor=east}}
        \tikzset{diagonal label/.style={rotate=45,anchor=north east}}
        \foreach \y in {1,...,\numClasses}
        {
            \node [anchor=east] at (0.4,-\y) {\pgfmathparse{\classNames[\y-1]}\pgfmathresult}; 
            
            \foreach \x in {1,...,\numClasses}
            {
            \def\totSamples{0}
            \foreach \ll in {1,...,\numClasses}
            {
                \pgfmathparse{\myConfMat[\ll-1][\x-1]} 
                \xdef\totSamples{\totSamples+\pgfmathresult}
            }
            \pgfmathparse{\totSamples} \xdef\totSamples{\pgfmathresult} 
            \begin{scope}[shift={(\x,-\y)}]
                \def\mVal{\myConfMat[\y-1][\x-1]} 
                \pgfmathtruncatemacro{\r}{\mVal}  
        \pgfmathtruncatemacro{\p}{round(\r/(0.0001+\totSamples)*100)}
                \coordinate (C) at (0,0);
                \ifthenelse{\p<50}{\def\txtcol{black}}{\def\txtcol{white}} 
                \node[
                    draw,   
                    text=\txtcol, 
                    align=center, 
                    fill=black!\p,  
                    minimum size=\myScale*10mm,
                    inner sep=0, 
                    ] (C) {\r\\\p\%};
                \ifthenelse{\y=\numClasses}{
                \node [] at ($(C)-(0,0.75)$)
                {\pgfmathparse{\classNames[\x-1]}\pgfmathresult};}{}
            \end{scope}
            }
        }
        \coordinate (yaxis) at (-0.3,0.5-\numClasses/2);
        \coordinate (xaxis) at (0.5+\numClasses/2, -\numClasses-1.25);
        \node [vertical label] at (yaxis) {Expert 1};
        \node []               at (xaxis) {Scale AI};
        \end{tikzpicture}
        
        \def\myConfMat{{
        {8,	0,	1,	0,	1,	0,	0},  
        {0,	10,	0,	0,	0,	0,	0},  
        {8,	57,	28,	7,	0,	0,	0},  
        {4,	13,	33,	35,	9,	3,	1},  
        {1,	1,	3,	6,	18,	16,	2},  
        {1,	0,	0,	0,	0,	19,	24},  
        {1,	0,	0,	0,	0,	0,	12},  
        }}
        \def\classNames{{"NA","1","2","3","4","5","6"}} 
        \def\numClasses{7}
        \def\myScale{0.7}
        \begin{tikzpicture}[
            scale = \myScale,
            font={\footnotesize}, 
            ]
        \tikzset{vertical label/.style={rotate=90,anchor=east}}
        \tikzset{diagonal label/.style={rotate=45,anchor=north east}}
        \foreach \y in {1,...,\numClasses}
        {
            \node [anchor=east] at (0.4,-\y) {\pgfmathparse{\classNames[\y-1]}\pgfmathresult}; 
            
            \foreach \x in {1,...,\numClasses}
            {
            \def\totSamples{0}
            \foreach \ll in {1,...,\numClasses}
            {
                \pgfmathparse{\myConfMat[\ll-1][\x-1]} 
                \xdef\totSamples{\totSamples+\pgfmathresult}
            }
            \pgfmathparse{\totSamples} \xdef\totSamples{\pgfmathresult} 
            \begin{scope}[shift={(\x,-\y)}]
                \def\mVal{\myConfMat[\y-1][\x-1]} 
                \pgfmathtruncatemacro{\r}{\mVal}  
        \pgfmathtruncatemacro{\p}{round(\r/(0.0001+\totSamples)*100)}
                \coordinate (C) at (0,0);
                \ifthenelse{\p<50}{\def\txtcol{black}}{\def\txtcol{white}} 
                \node[
                    draw,   
                    text=\txtcol, 
                    align=center, 
                    fill=black!\p,  
                    minimum size=\myScale*10mm,
                    inner sep=0, 
                    ] (C) {\r\\\p\%};
                \ifthenelse{\y=\numClasses}{
                \node [] at ($(C)-(0,0.75)$)
                {\pgfmathparse{\classNames[\x-1]}\pgfmathresult};}{}
            \end{scope}
            }
        }
        \coordinate (yaxis) at (-0.3,0.5-\numClasses/2);
        \coordinate (xaxis) at (0.5+\numClasses/2, -\numClasses-1.25);
        \node [vertical label] at (yaxis) {Expert 1};
        \node []               at (xaxis) {Centaur Labs};
        \end{tikzpicture}
    \end{multicols}
    
    \begin{multicols}{2}
        \def\myConfMat{{
        {0,	0,	1,	2,	0,	4,	3},  
        {0,	3,	6,	1,	0,	0,	0},  
        {0,	24,	46,	16,	0,	6,	8},  
        {0,	7,	31,	28,	5,	20,	7},  
        {0,	1,	8,	12,	2,	23,	1},  
        {0,	1,	2,	6,	1,	24,	10},  
        {0,	0,	0,	1,	0,	2,	10},  
        }}
        \def\classNames{{"NA","1","2","3","4","5","6"}} 
        \def\numClasses{7}
        \def\myScale{0.7}
        \begin{tikzpicture}[
            scale = \myScale,
            font={\footnotesize}, 
            ]
        \tikzset{vertical label/.style={rotate=90,anchor=east}}
        \tikzset{diagonal label/.style={rotate=45,anchor=north east}}
        \foreach \y in {1,...,\numClasses}
        {
            \node [anchor=east] at (0.4,-\y) {\pgfmathparse{\classNames[\y-1]}\pgfmathresult}; 
            
            \foreach \x in {1,...,\numClasses}
            {
            \def\totSamples{0}
            \foreach \ll in {1,...,\numClasses}
            {
                \pgfmathparse{\myConfMat[\ll-1][\x-1]} 
                \xdef\totSamples{\totSamples+\pgfmathresult}
            }
            \pgfmathparse{\totSamples} \xdef\totSamples{\pgfmathresult} 
            \begin{scope}[shift={(\x,-\y)}]
                \def\mVal{\myConfMat[\y-1][\x-1]} 
                \pgfmathtruncatemacro{\r}{\mVal}  
        \pgfmathtruncatemacro{\p}{round(\r/(0.0001+\totSamples)*100)}
                \coordinate (C) at (0,0);
                \ifthenelse{\p<50}{\def\txtcol{black}}{\def\txtcol{white}} 
                \node[
                    draw,   
                    text=\txtcol, 
                    align=center, 
                    fill=black!\p,  
                    minimum size=\myScale*10mm,
                    inner sep=0, 
                    ] (C) {\r\\\p\%};
                \ifthenelse{\y=\numClasses}{
                \node [] at ($(C)-(0,0.75)$)
                {\pgfmathparse{\classNames[\x-1]}\pgfmathresult};}{}
            \end{scope}
            }
        }
        \coordinate (yaxis) at (-0.3,0.5-\numClasses/2);
        \coordinate (xaxis) at (0.5+\numClasses/2, -\numClasses-1.25);
        \node [vertical label] at (yaxis) {Expert 1};
        \node []               at (xaxis) {Individual Typology Angle};
        \end{tikzpicture}
        
        \def\myConfMat{{
        {0,	0,	0,	0,	0,	0,	0},  
        {0,	13,	3,	0,	0,	0,	0},  
        {2,	17,	62,	6,	1,	0,	0},  
        {1,	9,	52,	20,	10,	2,	0},  
        {1,	1,	7,	8,	17,	12,	0},  
        {0,	0,	1,	2,	7,	27,	11},  
        {0,	0,	0,	0,	1,	7,	22},  
        }}
        \def\classNames{{"NA","1","2","3","4","5","6"}} 
        \def\numClasses{7}
        \def\myScale{0.7}
        \begin{tikzpicture}[
            scale = \myScale,
            font={\footnotesize}, 
            ]
        \tikzset{vertical label/.style={rotate=90,anchor=east}}
        \tikzset{diagonal label/.style={rotate=45,anchor=north east}}
        \foreach \y in {1,...,\numClasses}
        {
            \node [anchor=east] at (0.4,-\y) {\pgfmathparse{\classNames[\y-1]}\pgfmathresult}; 
            
            \foreach \x in {1,...,\numClasses}
            {
            \def\totSamples{0}
            \foreach \ll in {1,...,\numClasses}
            {
                \pgfmathparse{\myConfMat[\ll-1][\x-1]} 
                \xdef\totSamples{\totSamples+\pgfmathresult}
            }
            \pgfmathparse{\totSamples} \xdef\totSamples{\pgfmathresult} 
            \begin{scope}[shift={(\x,-\y)}]
                \def\mVal{\myConfMat[\y-1][\x-1]} 
                \pgfmathtruncatemacro{\r}{\mVal}  
        \pgfmathtruncatemacro{\p}{round(\r/(0.0001+\totSamples)*100)}
                \coordinate (C) at (0,0);
                \ifthenelse{\p<50}{\def\txtcol{black}}{\def\txtcol{white}} 
                \node[
                    draw,   
                    text=\txtcol, 
                    align=center, 
                    fill=black!\p,  
                    minimum size=\myScale*10mm,
                    inner sep=0, 
                    ] (C) {\r\\\p\%};
                \ifthenelse{\y=\numClasses}{
                \node [] at ($(C)-(0,0.75)$)
                {\pgfmathparse{\classNames[\x-1]}\pgfmathresult};}{}
            \end{scope}
            }
        }
        \coordinate (yaxis) at (-0.3,0.5-\numClasses/2);
        \coordinate (xaxis) at (0.5+\numClasses/2, -\numClasses-1.25);
        \node [vertical label] at (yaxis) {Expert 2};
        \node []               at (xaxis) {Scale AI};
        \end{tikzpicture}
    \end{multicols}

    \begin{multicols}{2}
        \def\myConfMat{{
        {0,	0,	0,	0,	0,	0,	0},  
        {2,	13,	1,	0,	0,	0,	0},  
        {8,	40,	29,	8,	3,	0,	0},  
        {9,	25,	28,	22,	9,	1,	0},  
        {1,	3,	5,	14,	12,	10,	1},  
        {1,	0,	2,	4,	4,	23,	14},  
        {2,	0,	0,	0,	0,	4,	24},  
        }}
        \def\classNames{{"NA","1","2","3","4","5","6"}} 
        \def\numClasses{7}
        \def\myScale{0.7}
        \begin{tikzpicture}[
            scale = \myScale,
            font={\footnotesize}, 
            ]
        \tikzset{vertical label/.style={rotate=90,anchor=east}}
        \tikzset{diagonal label/.style={rotate=45,anchor=north east}}
        \foreach \y in {1,...,\numClasses}
        {
            \node [anchor=east] at (0.4,-\y) {\pgfmathparse{\classNames[\y-1]}\pgfmathresult}; 
            
            \foreach \x in {1,...,\numClasses}
            {
            \def\totSamples{0}
            \foreach \ll in {1,...,\numClasses}
            {
                \pgfmathparse{\myConfMat[\ll-1][\x-1]} 
                \xdef\totSamples{\totSamples+\pgfmathresult}
            }
            \pgfmathparse{\totSamples} \xdef\totSamples{\pgfmathresult} 
            \begin{scope}[shift={(\x,-\y)}]
                \def\mVal{\myConfMat[\y-1][\x-1]} 
                \pgfmathtruncatemacro{\r}{\mVal}  
        \pgfmathtruncatemacro{\p}{round(\r/(0.0001+\totSamples)*100)}
                \coordinate (C) at (0,0);
                \ifthenelse{\p<50}{\def\txtcol{black}}{\def\txtcol{white}} 
                \node[
                    draw,   
                    text=\txtcol, 
                    align=center, 
                    fill=black!\p,  
                    minimum size=\myScale*10mm,
                    inner sep=0, 
                    ] (C) {\r\\\p\%};
                \ifthenelse{\y=\numClasses}{
                \node [] at ($(C)-(0,0.75)$)
                {\pgfmathparse{\classNames[\x-1]}\pgfmathresult};}{}
            \end{scope}
            }
        }
        \coordinate (yaxis) at (-0.3,0.5-\numClasses/2);
        \coordinate (xaxis) at (0.5+\numClasses/2, -\numClasses-1.25);
        \node [vertical label] at (yaxis) {Expert 2};
        \node []               at (xaxis) {Centaur Labs};
        \end{tikzpicture}
        
        \def\myConfMat{{
        {0,	0,	0,	0,	0,	0,	0},  
        {0,	5,	8,	0,	0,	2,	1},  
        {0,	13,	40,	18,	2,	6,	9},  
        {0,	15,	32,	26,	2,	16,	3},  
        {0,	2,	8,	11,	3,	19,	3},  
        {0,	1,	6,	9,	0,	25,	7},  
        {0, 0,	0,	2,	1,	11,	16},  
        }}
        \def\classNames{{"NA","1","2","3","4","5","6"}} 
        \def\numClasses{7}
        \def\myScale{0.7}
        \begin{tikzpicture}[
            scale = \myScale,
            font={\footnotesize}, 
            ]
        \tikzset{vertical label/.style={rotate=90,anchor=east}}
        \tikzset{diagonal label/.style={rotate=45,anchor=north east}}
        \foreach \y in {1,...,\numClasses}
        {
            \node [anchor=east] at (0.4,-\y) {\pgfmathparse{\classNames[\y-1]}\pgfmathresult}; 
            
            \foreach \x in {1,...,\numClasses}
            {
            \def\totSamples{0}
            \foreach \ll in {1,...,\numClasses}
            {
                \pgfmathparse{\myConfMat[\ll-1][\x-1]} 
                \xdef\totSamples{\totSamples+\pgfmathresult}
            }
            \pgfmathparse{\totSamples} \xdef\totSamples{\pgfmathresult} 
            \begin{scope}[shift={(\x,-\y)}]
                \def\mVal{\myConfMat[\y-1][\x-1]} 
                \pgfmathtruncatemacro{\r}{\mVal}  
        \pgfmathtruncatemacro{\p}{round(\r/(0.0001+\totSamples)*100)}
                \coordinate (C) at (0,0);
                \ifthenelse{\p<50}{\def\txtcol{black}}{\def\txtcol{white}} 
                \node[
                    draw,   
                    text=\txtcol, 
                    align=center, 
                    fill=black!\p,  
                    minimum size=\myScale*10mm,
                    inner sep=0, 
                    ] (C) {\r\\\p\%};
                \ifthenelse{\y=\numClasses}{
                \node [] at ($(C)-(0,0.75)$)
                {\pgfmathparse{\classNames[\x-1]}\pgfmathresult};}{}
            \end{scope}
            }
        }
        \coordinate (yaxis) at (-0.3,0.5-\numClasses/2);
        \coordinate (xaxis) at (0.5+\numClasses/2, -\numClasses-1.25);
        \node [vertical label] at (yaxis) {Expert 2};
        \node []               at (xaxis) {Individual Typology Angle};
        \end{tikzpicture}
    \end{multicols}
    
    \caption{Confusion matrices comparing experts 1 and 2 to the ITA algorithm and crowdsourcing methods.}
    \label{fig:cf_matrices}
\end{figure}

\begin{figure}[p]

    \begin{multicols}{2}
        \def\myConfMat{{
        {3,  1,  6,  0,  0,  0,  0},  
        {0,  9,  1,  0,  0,  0,  0},  
        {0,  25, 69, 5,  1,  0,  0},  
        {1,  5, 48, 24, 16,  4,  0},  
        {0,  0,  1,  7, 17, 21,  1},  
        {0,  0,  0,  0,  2, 21, 21},  
        {0,  0,  0,  0,  0,  2, 11},  
        }}
        \def\classNames{{"NA","1","2","3","4","5","6"}} 
        \def\numClasses{7}
        \def\myScale{0.7}
        \begin{tikzpicture}[
            scale = \myScale,
            font={\footnotesize}, 
            ]
        \tikzset{vertical label/.style={rotate=90,anchor=east}}
        \tikzset{diagonal label/.style={rotate=45,anchor=north east}}
        \foreach \y in {1,...,\numClasses}
        {
            \node [anchor=east] at (0.4,-\y) {\pgfmathparse{\classNames[\y-1]}\pgfmathresult}; 
            
            \foreach \x in {1,...,\numClasses}
            {
            \def\totSamples{0}
            \foreach \ll in {1,...,\numClasses}
            {
                \pgfmathparse{\myConfMat[\ll-1][\x-1]} 
                \xdef\totSamples{\totSamples+\pgfmathresult}
            }
            \pgfmathparse{\totSamples} \xdef\totSamples{\pgfmathresult} 
            \begin{scope}[shift={(\x,-\y)}]
                \def\mVal{\myConfMat[\y-1][\x-1]} 
                \pgfmathtruncatemacro{\r}{\mVal}  
        \pgfmathtruncatemacro{\p}{round(\r/(0.0001+\totSamples)*100)}
                \coordinate (C) at (0,0);
                \ifthenelse{\p<50}{\def\txtcol{black}}{\def\txtcol{white}} 
                \node[
                    draw,   
                    text=\txtcol, 
                    align=center, 
                    fill=black!\p,  
                    minimum size=\myScale*10mm,
                    inner sep=0, 
                    ] (C) {\r\\\p\%};
                \ifthenelse{\y=\numClasses}{
                \node [] at ($(C)-(0,0.75)$)
                {\pgfmathparse{\classNames[\x-1]}\pgfmathresult};}{}
            \end{scope}
            }
        }
        \coordinate (yaxis) at (-0.3,0.5-\numClasses/2);
        \coordinate (xaxis) at (0.5+\numClasses/2, -\numClasses-1.25);
        \node [vertical label] at (yaxis) {Expert 3};
        \node []               at (xaxis) {Scale AI};
        \end{tikzpicture}
        
        \def\myConfMat{{
        {3,	1,	2,	0,	1,	1,	0},  
        {0,	6,	4,	0,	0,	0,	0},  
        {0,	30,	86,	15,	3,	1,	0},  
        {0,	3,	27,	11,	6,	0,	0},  
        {1,	0,	5,	6,	21,	6,	0},  
        {0,	0,	1,	3,	6,	32,	6},  
        {0,	0,	0,	0,	0,	7,	27},  
        }}
        \def\classNames{{"NA","1","2","3","4","5","6"}} 
        \def\numClasses{7}
        \def\myScale{0.7}
        \begin{tikzpicture}[
            scale = \myScale,
            font={\footnotesize}, 
            ]
        \tikzset{vertical label/.style={rotate=90,anchor=east}}
        \tikzset{diagonal label/.style={rotate=45,anchor=north east}}
        \foreach \y in {1,...,\numClasses}
        {
            \node [anchor=east] at (0.4,-\y) {\pgfmathparse{\classNames[\y-1]}\pgfmathresult}; 
            
            \foreach \x in {1,...,\numClasses}
            {
            \def\totSamples{0}
            \foreach \ll in {1,...,\numClasses}
            {
                \pgfmathparse{\myConfMat[\ll-1][\x-1]} 
                \xdef\totSamples{\totSamples+\pgfmathresult}
            }
            \pgfmathparse{\totSamples} \xdef\totSamples{\pgfmathresult} 
            \begin{scope}[shift={(\x,-\y)}]
                \def\mVal{\myConfMat[\y-1][\x-1]} 
                \pgfmathtruncatemacro{\r}{\mVal}  
        \pgfmathtruncatemacro{\p}{round(\r/(0.0001+\totSamples)*100)}
                \coordinate (C) at (0,0);
                \ifthenelse{\p<50}{\def\txtcol{black}}{\def\txtcol{white}} 
                \node[
                    draw,   
                    text=\txtcol, 
                    align=center, 
                    fill=black!\p,  
                    minimum size=\myScale*10mm,
                    inner sep=0, 
                    ] (C) {\r\\\p\%};
                \ifthenelse{\y=\numClasses}{
                \node [] at ($(C)-(0,0.75)$)
                {\pgfmathparse{\classNames[\x-1]}\pgfmathresult};}{}
            \end{scope}
            }
        }
        \coordinate (yaxis) at (-0.3,0.5-\numClasses/2);
        \coordinate (xaxis) at (0.5+\numClasses/2, -\numClasses-1.25);
        \node [vertical label] at (yaxis) {Expert 3};
        \node []               at (xaxis) {Centaur Labs};
        \end{tikzpicture}
    \end{multicols}
    
      \begin{multicols}{2}
        \def\myConfMat{{
        {6,  0,  0,  1,  0,  0,  0},  
        {2,  7,  1,  0,  0,  0,  0},  
        {8,  65, 38, 21,  2,  0,  1},  
        {4,  9, 19, 11, 4,  0,  0},  
        {1,  0,  4,  12, 15,7,  0},  
        {0,  0,  3,  2,  7, 29, 7},  
        {2,  0,  0,  0,  0,  2, 30},  
        }}
        \def\classNames{{"NA","1","2","3","4","5","6"}} 
        \def\numClasses{7}
        \def\myScale{0.7}
        \begin{tikzpicture}[
            scale = \myScale,
            font={\footnotesize}, 
            ]
        \tikzset{vertical label/.style={rotate=90,anchor=east}}
        \tikzset{diagonal label/.style={rotate=45,anchor=north east}}
        \foreach \y in {1,...,\numClasses}
        {
            \node [anchor=east] at (0.4,-\y) {\pgfmathparse{\classNames[\y-1]}\pgfmathresult}; 
            
            \foreach \x in {1,...,\numClasses}
            {
            \def\totSamples{0}
            \foreach \ll in {1,...,\numClasses}
            {
                \pgfmathparse{\myConfMat[\ll-1][\x-1]} 
                \xdef\totSamples{\totSamples+\pgfmathresult}
            }
            \pgfmathparse{\totSamples} \xdef\totSamples{\pgfmathresult} 
            \begin{scope}[shift={(\x,-\y)}]
                \def\mVal{\myConfMat[\y-1][\x-1]} 
                \pgfmathtruncatemacro{\r}{\mVal}  
        \pgfmathtruncatemacro{\p}{round(\r/(0.0001+\totSamples)*100)}
                \coordinate (C) at (0,0);
                \ifthenelse{\p<50}{\def\txtcol{black}}{\def\txtcol{white}} 
                \node[
                    draw,   
                    text=\txtcol, 
                    align=center, 
                    fill=black!\p,  
                    minimum size=\myScale*10mm,
                    inner sep=0, 
                    ] (C) {\r\\\p\%};
                \ifthenelse{\y=\numClasses}{
                \node [] at ($(C)-(0,0.75)$)
                {\pgfmathparse{\classNames[\x-1]}\pgfmathresult};}{}
            \end{scope}
            }
        }
        \coordinate (yaxis) at (-0.3,0.5-\numClasses/2);
        \coordinate (xaxis) at (0.5+\numClasses/2, -\numClasses-1.25);
        \node [vertical label] at (yaxis) {Expert 3};
        \node []               at (xaxis) {Individual Typology Angle};
        \end{tikzpicture}
        
        \def\myConfMat{{
        {0,	0,	1,	1,	0,	1,	4},  
        {0,	2,	6,	0,	0,	1,	1},  
        {0,	26,	62,	25,	2,	12,	8},  
        {0,	4,	11,	19,	2,	6,	5},  
        {0,	2,	6,	9,	3,	18,	1},  
        {0,	2,	8,	8,	0,	26,	4},  
        {0,	0,	0,	3,	1,	14,	16},  
        }}
    \end{multicols}
    \caption{Confusion matrices comparing expert 3 to the ITA algorithm and crowdsourcing methods.}
    \label{fig:cf_matrices_3}
\end{figure}
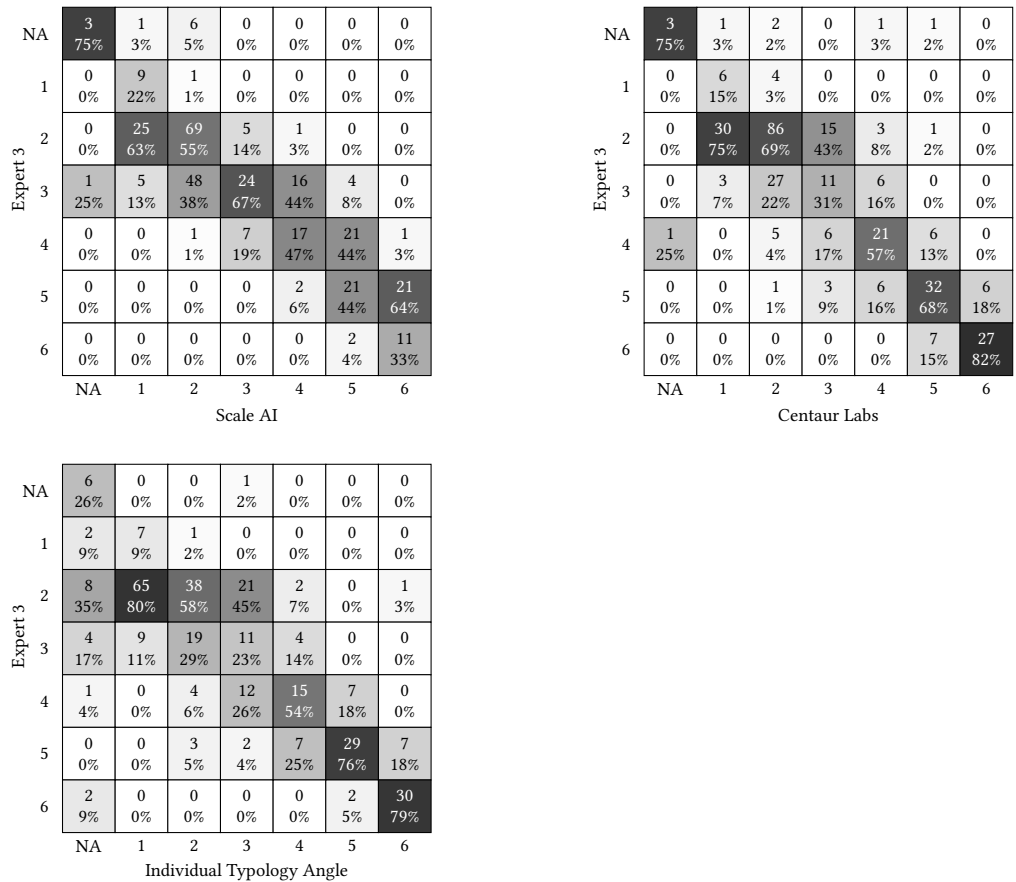

\end{document}